\documentclass[final,authoryear,5p,times,twocolumn]{elsarticle}
\pdfoutput=1

\usepackage{graphicx}
\usepackage{subfig}
\usepackage[cmex10]{amsmath}
\usepackage{amssymb}
\usepackage{multirow}
\usepackage{url}
\usepackage{placeins}
\usepackage{floatrow}
\usepackage{color}
\usepackage{flushend}

\usepackage{tabularx}
\usepackage{array}
\usepackage{rotating}

\definecolor{dark-gray}{gray}{0.5}
\definecolor{dark-green}{rgb}{0, 0.4, 0}

\newcolumntype{Y}{>{\centering\arraybackslash}X}
\newcommand\RotText[1]{\rotatebox{90}{\parbox{2.2cm}{\centering#1}}}

\journal{Medical Image Analysis}

\begin{document}

\begin{frontmatter}



\title{Automatic detection of rare pathologies in fundus photographs using few-shot learning}

\author[label1]{Gwenol\'e~Quellec\corref{cor1}}
\ead{gwenole.quellec@inserm.fr}
\author[label2,label1]{Mathieu~Lamard}
\author[label3,label1]{Pierre-Henri~Conze}
\author[label4]{Pascale~Massin}
\author[label2,label1,label5]{B\'eatrice~Cochener}
\cortext[cor1]{LaTIM - IBRBS - 22, avenue Camille Desmoulins - 29200 Brest, France - Tel.: +33 2 98 01 81 29}
\address[label1]{Inserm, UMR 1101, Brest, F-29200 France}
\address[label2]{Univ Bretagne Occidentale, Brest, F-29200 France}
\address[label3]{IMT Atlantique, Brest, F-29200 France}
\address[label4]{Service d'Ophtalmologie, H\^{o}pital Lariboisi\`ere, APHP, Paris, F-75475 France}
\address[label5]{Service d'Ophtalmologie, CHRU Brest, Brest, F-29200 France}

\begin{abstract}
In the last decades, large datasets of fundus photographs have been collected in diabetic retinopathy (DR) screening networks. Through deep learning, these datasets were used to train automatic detectors for DR and a few other frequent pathologies, with the goal to automate screening. One challenge limits the adoption of such systems so far: automatic detectors ignore rare conditions that ophthalmologists currently detect, such as papilledema or anterior ischemic optic neuropathy. The reason is that standard deep learning requires too many examples of these conditions. However, this limitation can be addressed with few-shot learning, a machine learning paradigm where a classifier has to generalize to a new category not seen in training, given only a few examples of this category. This paper presents a new few-shot learning framework that extends convolutional neural networks (CNNs), trained for frequent conditions, with an unsupervised probabilistic model for rare condition detection. It is based on the observation that CNNs often perceive photographs containing the same anomalies as similar, even though these CNNs were trained to detect unrelated conditions. This observation was based on the t-SNE visualization tool, which we decided to incorporate in our probabilistic model. Experiments on a dataset of 164,660 screening examinations from the OPHDIAT screening network show that 37 conditions, out of 41, can be detected with an area under the ROC curve (AUC) greater than 0.8 (average AUC: 0.938). In particular, this framework significantly outperforms other frameworks for detecting rare conditions, including multitask learning, transfer learning and Siamese networks, another few-shot learning solution. We expect these richer predictions to trigger the adoption of automated eye pathology screening, which will revolutionize clinical practice in ophthalmology.
\end{abstract}


\end{frontmatter}

\section{Introduction}
\label{sec:Introduction}

According to the World Health Organization, 285 million people are visually impaired worldwide, but the preventable causes represent 80\% of the total burden \citep{pascolini_global_2012}. Early detection and management of ocular pathologies is one major strategy to prevent vision impairment. With the recent success of deep learning, many automatic screening systems based on fundus photography were proposed recently. Diabetic retinopathy (DR) was historically the first pathology targeted by those systems \citep{gulshan_development_2016, abramoff_improved_2016, QuellecDeepimagemining2017a, raju_development_2017, gargeya_automated_2017, QuellecInstantautomaticdiagnosis2019, nielsen_deep_2019}. The reason is that large datasets of images have been collected within DR screening programs for diabetic patients over the past decades \citep{massin_ophdiat:_2008, cuadros_eyepacs:_2009}: those images were interpreted by human readers, which allows efficient training of supervised deep learning classifiers. Automatic screening systems were also proposed for glaucoma \citep{li_efficacy_2018, shibata_development_2018, christopher_performance_2018, phan_evaluation_2019, ahn_deep_2019, diaz-pinto_cnns_2019} and age-related macular degeneration (AMD) \citep{matsuba_accuracy_2018, pead_automated_nodate}, the other two major sight-threatening pathologies in developed countries. Other pathologies such as retinopathy of prematurity \citep{wang_automated_2018} have also been targeted. A few studies also addressed multiple pathology screening \citep{keel_visualizing_2018, choi_multi-categorical_2017}. \citet{ting_development_2017} thus proposed to detect AMD and glaucoma, in addition to DR, in DR screening images. The motivation is that diabetic patients, targeted by DR screening programs, may also suffer from AMD or glaucoma: ophthalmologists may not be willing to replace their interpretations with automatic interpretations if other sight-threatening pathologies are ignored.

In this study, we propose to go one step further and detect all conditions annotated by human readers in DR screening reports. In the OPHDIAT screening program \citep{massin_ophdiat:_2008}, for instance, this represents 41 conditions. Targeting those conditions has become possible because more than 160,000 screening examinations ($>$ 760,000 images) have been performed so far. Yet some of these conditions are still very rare and appear in less than ten screening reports: this impacts the type of machine learning (ML) strategy to employ. In particular, training a regular deep learning model for each of these conditions is prohibitive, even through transfer learning \citep{CheplyginaCatsCATscans2019a}. Targeting 41 conditions is a big leap compared to the state of the art. \citet{choi_multi-categorical_2017} focused on the classification of 10 pathologies, but not in a screening context: the goal was to differentiate pathologies, not to detect them in a large population. \citet{fauw_clinically_2018} mention that they target 53 ``key diagnoses'' in optical coherence tomography, but these diagnoses are not listed and the detection performance not reported: the main goal was to propose automatic referral decisions. We expect this additional information to facilitate the adoption of automatic screening.

This paper presents the ML solution we propose to address the challenge of detecting rare conditions. The genesis of this framework was the use of t-distributed stochastic neighbor embedding (t-SNE) \citep{vanderMaatenVisualizinghighdimensionaldata2008} to visualize what convolutional neural networks (CNNs), trained to detect DR on the OPHDIAT dataset \citep{QuellecInstantautomaticdiagnosis2019}, have learnt. We observed that many conditions unrelated to DR were clustered in feature space, even though the models were only trained to detect DR. This suggests that CNNs are performing differential diagnosis to detect DR. We hypothesized that CNNs trained to detect several frequent conditions simultaneously could improve this phenomenon further. Therefore, in the proposed framework, a standard deep learning classifier is trained to detect frequent conditions and simple probabilistic models are derived from these deep learning models to detect rare conditions. As such, this framework solves a few-shot learning problem, a ML paradigm where a classifier must generalize to a new category not seen in training, given only a few examples of this category \citep{WangGeneralizingfewexamples2019}. One specificity of this framework is that probabilistic models rely on standard classification CNNs and on t-SNE. It combines ideas from transfer learning and multitask learning, while outperforming each of these frameworks individually, as demonstrated in this paper.

The paper is organized as follows. Related ML frameworks are presented in Section \ref{sec:RelatedMachineLearningFrameworks}. The proposed framework is described in Section \ref{sec:SpinoffLearning}. Experiments in the OPHDIAT dataset are reported in Section \ref{sec:Experiments}. We end up with a discussion and conclusions in Section \ref{sec:DiscussionConclusions}.

\section{Related Machine Learning Frameworks}
\label{sec:RelatedMachineLearningFrameworks}

A well-known solution for dealing with data scarcity is \textit{transfer learning} \citep{CheplyginaCatsCATscans2019a}. In transfer learning, an initial classification model is trained on a large dataset, such as ImageNet (1.2 million images)\footnote{\url{http://www.image-net.org}}, to perform unrelated tasks. Then, this model is fine-tuned on the dataset of interest, to detect a target condition. The idea is that parts of the feature extraction process, such as edge detection, are common to many computer vision tasks and can therefore be reused, with or without modifications. This approach has become the leading strategy in medical image analysis \citep{Litjenssurveydeeplearning2017a}. Another solution to this problem is \textit{multitask learning} \citep{CaruanaMultitasklearning1997}. The difference with transfer learning is that one learns to address multiple tasks simultaneously rather than sequentially. In multitask learning, auxiliary tasks are usually chosen because training labels are abundant or not needed, unlike the target task \citep{ZhangFaciallandmarkdetection2014, MordanRevisitingMultiTaskLearning2018}. Multitask learning can thus be used to train a unique detector for multiple (both rare and frequent) conditions \citep{GuendelMultitasklearningchest2019}: detecting frequent conditions can be regarded as an auxiliary task, for the main task of detecting rare conditions.

A more recent solution to this problem is \textit{one-shot learning} \citep{Fei-FeiOneshotlearningobject2006} or more generally \textit{few-shot learning} \citep{WangGeneralizingfewexamples2019}. In one-shot or few-shot learning, a classifier must generalize to a new category not seen in training, given only one or a few examples of this category. In the context of deep learning, one increasingly popular solution is to design a neural network accepting two images as inputs and deciding whether or not these images belong to the same category. Such networks include Siamese networks \citep{KochSiameseneuralnetworks2015, ShyamAttentiverecurrentcomparators2017}, matching networks \citep{VinyalsMatchingnetworksone2016} or relation networks \citep{SungLearningcompareRelation2018}. Another solution to this problem is to design a simple probabilistic model for the new category: this model operates in an image feature space derived from the initial training \citep{Fei-FeiOneshotlearningobject2006}. \citet{Fei-FeiOneshotlearningobject2006} applied this strategy to local features from the pre-deep-learning era. \citet{SnellPrototypicalnetworksfewshot2017} applied this strategy to an image feature space derived from matching networks \citep{VinyalsMatchingnetworksone2016}. This strategy was also applied in this paper. However, we propose that the image feature space derives from usual (single-image) classification CNNs. Therefore, we can take full advantage of the large literature on classification CNNs. We can also take advantage of pre-trained models available for these CNNs. As explained hereafter, the key to the proposed solution is the use of t-SNE in the design of an image feature space.

\section{Proposed Few-Shot Learning Framework}
\label{sec:SpinoffLearning}

\begin{figure*}[!t]
  \begin{center}
    \includegraphics[width=\textwidth]{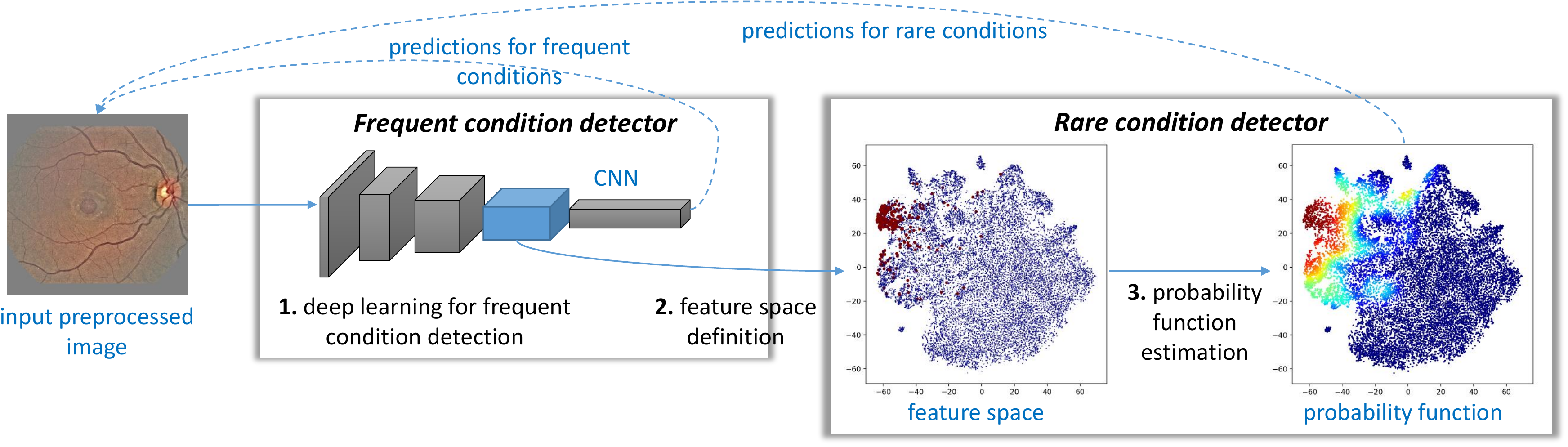}
  \end{center}
  \caption{Proposed pipeline (learning and inference). To initiate the learning phase, a CNN is trained to detect frequent conditions in preprocessed images (\textbf{1.}). This CNN is then used to build detectors for rare conditions. In that purpose, a feature space is designed: features derive from the output of selected neurons of the CNN (\textbf{2.}). Next, a probabilistic model is trained for each rare condition in this feature space (\textbf{3.}). The inference phase is similar: preprocessed images are processed by the CNN, and a prediction is made for both frequent and rare conditions. Predictions for frequent conditions simply are the CNN outputs. Predictions for rare condition are further inferred by the probabilistic models.}
  \label{fig:pipeline}
\end{figure*}

The proposed framework, illustrated in Fig.~\ref{fig:pipeline}, can be summarized as follows. A multitask detector for frequent conditions is trained first (see Section \ref{sec:DeepLearningFrequentConditionDetection}). Next, a probabilistic detection model is defined for each rare condition (see Sections \ref{sec:FeatureSpaceDefinition} to \ref{sec:ProbabilityFunctionEstimation}). Then, predictions can be inferred for new images: predictions are computed for both frequent and rare conditions (see Section \ref{sec:RareConditionDetection}).

\subsection{Notations}
\label{sec:Notations}

Let $\mathcal{D}$ denote an image dataset where the presence or absence of $N$ conditions has been annotated by one or multiple human readers for each image $I \in \mathcal{D}$.  Let $(c_n)_{n=1..N}$ denote these conditions. Let $y_{I,n} \in \{0,1\}$ denote a label indicating the presence ($y_{I,n}=1$) or absence ($y_{I,n}=0$) of condition $c_n$ in image $I$ according to experts. Let $f_n$ denote the frequency (the raw count) of condition $c_n$ in dataset $\mathcal{D}$: $f_n=\sum_{I \in \mathcal{D}}{y_{I,n}}$. Conditions are sorted by decreasing frequency order: $f_{n'}\leq f_n, \forall n'\geq n$.

We assume that dataset $\mathcal{D}$ was divided into a learning (or training) subset $\mathcal{D}_L$, used for deep learning, a validation subset $\mathcal{D}_V$, and a test subset $\mathcal{D}_T$. These datasets are mutually exclusive ($\mathcal{D}_L \cap \mathcal{D}_V = \mathcal{D}_L \cap \mathcal{D}_T = \mathcal{D}_V \cap \mathcal{D}_T = \emptyset$, $\mathcal{D}_L \cup \mathcal{D}_V \cup \mathcal{D}_T = \mathcal{D}$). We also define a subset of ``reference images'' $\mathcal{D}_R$, whose definition will vary depending on whether the algorithm is being validated or tested (see Section \ref{sec:LearningValidationTesting}).

\begin{figure*}[!t]
  \begin{center}
    \includegraphics[width=.8\textwidth]{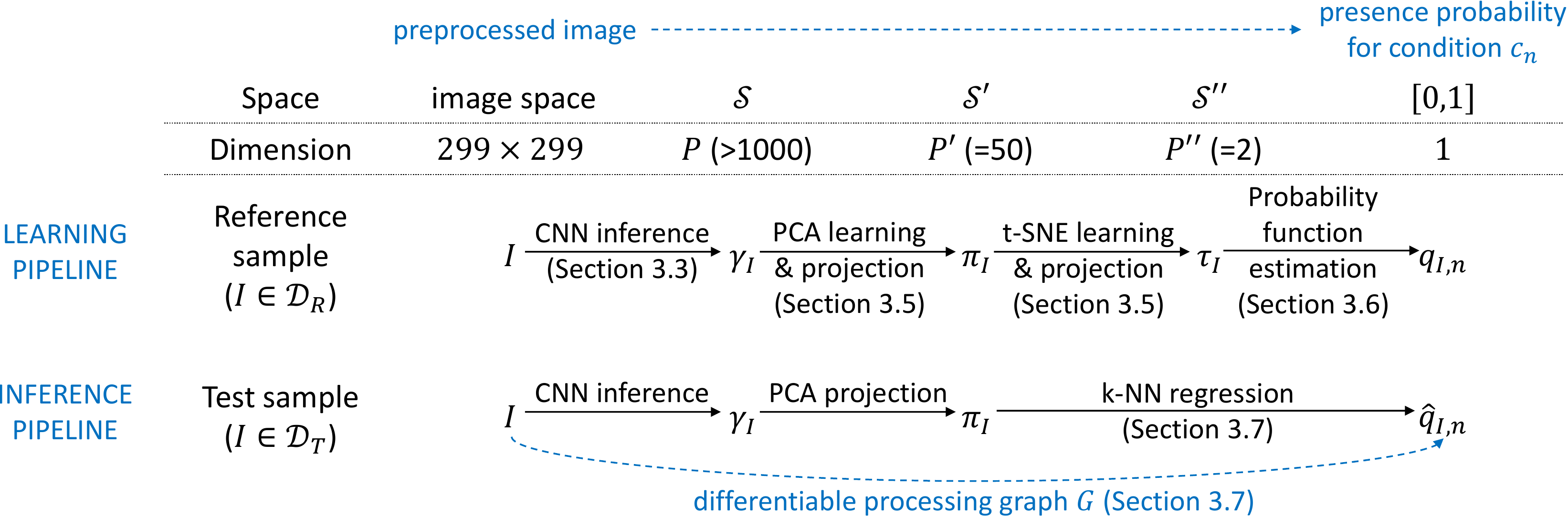}
  \end{center}
  \caption{Detailed pipeline for rare condition detection. This figure summarizes the intermediate steps for detecting a rare condition in images, as well as the associated notations. Two different pipelines are presented. The first pipeline is applied to reference images for learning a presence probability function (learning pipeline). The second pipeline is applied to test images for inferring presence probabilities (inference pipeline). Section \ref{sec:Experiments} explains how the value of each dimension was determined.}
  \label{fig:detailedPipeline}
\end{figure*}

All processing steps described hereafter are performed on preprocessed images (see Section \ref{sec:ImagePreprocessing}): for simplicity, $I$ denotes the preprocessed image in the following sections. Various spaces are defined hereafter to compute presence probabilities for rare conditions: those notations are summarized in Fig.~\ref{fig:detailedPipeline}.

\subsection{Deep Learning for Frequent Condition Detection}
\label{sec:DeepLearningFrequentConditionDetection}

The first step is to define a deep learning model for recognizing the $M \leq N$ most frequent conditions. This model relies on a convolutional neural network (CNN). This CNN is defined as a multilabel classifier; it is trained to minimize the following cost function $\mathcal{L}$:
\begin{equation}
  \left\lbrace
    \begin{array}{c}
      \mathcal{L} = -\displaystyle\sum_{n \leq M}\sum_{I \in \mathcal{D}_L}{ y_{I,n}\log\sigma(x_{I,n}) + (1 - y_{I,n})\log(1 - \sigma(x_{I,n})) } \\
      \sigma(x) = \displaystyle\frac{1}{1 + e^{-x}}
    \end{array}
  \right. \;,
\end{equation}
where $x_{I,n} \in \mathbb{R}$ denotes the output of the model for image $I$ and condition $c_n, n\leq M$. Through the logistic function $\sigma$, this output is converted into a probability $p_{I,n}^M = \sigma(x_{I,n}) \in \left[0,1\right]$ (simply noted $p_{I,n}$ in the absence of ambiguity). $\sigma$ was selected as activation function since patients can have multiple conditions simultaneously, which is properly modeled by multiple logistic functions. We note that training this initial classification model, defined for frequent conditions, is a multitask learning problem (see Section \ref{sec:RelatedMachineLearningFrameworks}): the proposed framework extends multitask learning to rare conditions as described hereafter.

\subsection{Feature Space Definition}
\label{sec:FeatureSpaceDefinition}

Since a unique CNN is defined to detect the $M$ most frequent conditions, the penultimate layer of this CNN is very general: it extracts all features required to detect $M$ conditions. We use the output of this layer to define a feature space in which the remaining $N - M$ conditions will be detected. Let $\mathcal{S}$ denote this feature space and  $\gamma_I$ the projection of a given image $I$ in this space (see Fig.~\ref{fig:detailedPipeline}). The number of neurons in the penultimate layer of a classification CNN is generally high, for instance 2,049 for Inception-v3 \citep{SzegedyRethinkinginceptionarchitecture2016} or 1,537 for inception-v4 \citep{SzegedyInceptionv4InceptionResNetimpact2017}: let $P$ denote the dimension of this space ($P$ = 2,049 or 1,537 for instance).

To address the curse of dimensionality, dimension reduction is performed afterwards. For this purpose, we propose the use of t-SNE, a nonlinear technique for embedding high-dimensional data in a low-dimensional space suited for visualization: typically a 2-D or 3-D space \citep{vanderMaatenVisualizinghighdimensionaldata2008}. In t-SNE, dimension reduction is unsupervised, but it is data-driven: it relies on the $\mathcal{D}_R$ reference subset.

\subsection{\textit{t}-distributed Stochastic Neighbor Embedding (t-SNE)}
\label{sec:tSNE}

In t-SNE, some high-dimensional input vectors $(\pi_I)_{I\in \mathcal{D}_R}$ are mapped to low-dimensional output vectors $(\tau_I)_{I\in \mathcal{D}_R}$ in such a way that similar input vectors are mapped to nearby output vectors and dissimilar input vectors are mapped to distant output vectors. First, t-SNE defines the conditional probability $p_{J\mid I}$ that sample $I$ picks sample $J$ as a neighbor. It assumes that neighbors are picked in proportion to their probability density under a Gaussian centered at $\pi_I$:
\begin{equation}
  p_{J\mid I} = \displaystyle\frac{\Gamma\left(\frac{\pi_I-\pi_J}{h_I}\right)}{\sum_{K\in \mathcal{D}_R\setminus \{I\}}{\Gamma\left(\frac{\pi_I-\pi_K}{h_I}\right)}} \;\;\; ,
  \label{eq:tSNE}
\end{equation}
where $\Gamma$ is a Gaussian kernel and $h_I$ is a sample-specific bandwidth:
\begin{equation}
  \Gamma(x) = \frac{1}{\sqrt{2\pi}}e^{-\frac{1}{2}x^2} \;\;\;.
  \label{eq:GaussianKernel}
\end{equation}
Let $P_I=(p_{J\mid I})_{J\in \mathcal{D}_R}$ denote the conditional distribution thus defined. Bandwidths are set in such a way that the perplexity $\rho(P_I)$ of $P_I$, interpreted by \citet{vanderMaatenVisualizinghighdimensionaldata2008} as a smooth measure of the effective number of neighbors of $\pi_I$, equals a predefined perplexity $\bar{\rho}$:
\begin{equation}
  \rho(P_I) = 2^{\displaystyle{-\sum_{J\in \mathcal{D}_R}{p_{J\mid I} \log_2 p_{J\mid I}}}} \;\;\;.
  \label{eq:Perplexity}
\end{equation}
Similar conditional distributions $Q_I=(q_{J\mid I})_{J\in \mathcal{D}_R}$ are defined for output vectors $(\tau_I)_{I\in \mathcal{D}_R}$, using a constant bandwidth $h=1/\sqrt{2}$. These output vectors can thus be found by minimizing the sum $C$ of Kullback-Leibler divergences between conditional distributions $P_I$ and $Q_I$, $I\in \mathcal{D}_R$, using a gradient descent:
\begin{equation}
  C = \sum_{I\in \mathcal{D}_R}\sum_{J\in \mathcal{D}_R}{p_{J\mid I}\log\frac{p_{J\mid I}}{q_{J\mid I}}} \;\;\;.
  \label{eq:Perplexity}
\end{equation}

\subsection{Feature Space Dimension Reduction}
\label{sec:FeatureSpaceDimensionReduction}

Following \citet{vanderMaatenVisualizinghighdimensionaldata2008}'s recommendation, a two-step procedure was in fact adopted for dimension reduction (see Fig.~\ref{fig:detailedPipeline}):
\begin{itemize}
	\item A first reduction step relies on principal component analysis (PCA), which transforms the initial feature space $\mathcal{S}$ into a new $P'$-dimensional feature space $\mathcal{S}'$. Let $\pi_I$ denote the projection of image $I$ into $\mathcal{S}'$: this will be the high-dimensional input vector of Section \ref{sec:tSNE}.
	\item In a second step, t-SNE itself transforms $\mathcal{S}'$ into a $P''$-dimensional feature space $\mathcal{S}''$. Let $\tau_I$ denote the projection of image $I$ into $\mathcal{S}''$: this is the low-dimensional output vector of Section \ref{sec:tSNE}.
\end{itemize}
As explained by \citet{vanderMaatenVisualizinghighdimensionaldata2008}, the use of PCA speeds up computation of pairwise distances between input vectors and it suppresses some noise without severely distorting the inter-sample distances.

\subsection{Probability Function Estimation}
\label{sec:ProbabilityFunctionEstimation}

As mentioned in the introduction, we observed that the t-SNE algorithm generates a feature space $\mathcal{S}''$ allowing very good separation of the various $(c_n)_{n=1..N}$ conditions, even though it is unsupervised. This observation is leveraged to define a probabilistic condition detection model in $\mathcal{S}''$ space. For that purpose, a density probability function $F_n$ is first defined in $\mathcal{S}''$ for each condition $c_n, n \leq N$. Probability density functions $\overline{F_n}$ are also defined for the absence of each condition. These estimations are performed in the $\mathcal{D}_R$ reference subset; training images are discarded in case the CNN has overfitted the training data. Density estimations rely on the Parzen-Rosenblatt method \citep{Parzenestimationprobabilitydensity1962}, using the Gaussian kernel $\Gamma$ of Equation (\ref{eq:GaussianKernel}). For each location $\tau \in \mathcal{S}''$:
\begin{equation}
  \left\lbrace
    \begin{array}{rcl}
      F_n(\tau)            &=& \displaystyle\frac{1}{ \sum_{I \in \mathcal{D}_R}{y_{I,n}} } \sum_{ I \in \mathcal{D}_R }{ \Gamma\left( \frac{ \tau - \tau_I }{h_n} \right) } \\
      \overline{F_n}(\tau) &=& \displaystyle\frac{1}{ \sum_{I \in \mathcal{D}_R}{[1 - y_{I,n}]} } \sum_{ I \in \mathcal{D}_R }{ \Gamma\left( \frac{ \tau - \tau_I }{\overline{h_n}} \right) }
    \end{array}
  \right. \;.
  \label{eq:probabilityDensityFunctions}
\end{equation}
For each density function, one parameter needs to be set: the $h_n$ or $\overline{h_n}$ bandwidth, which controls the smoothness of the estimated function. This parameter is set in an unsupervised fashion, according to Scott's criterion \citep{ScottMultivariatedensityestimation1992}:
\begin{equation}
  \begin{array}{cc}
    h_n = \displaystyle\left( \sum_{I \in \mathcal{D}_R}{y_{I,n}} \right)^{-\frac{1}{P'' + 4}}, &
    \overline{h_n} = \displaystyle\left( \sum_{I \in \mathcal{D}_R}{[1 - y_{I,n}]} \right)^{-\frac{1}{P'' + 4}}
  \end{array} \;.
\end{equation}
Finally, based on these two probability density functions, $F_n$ and $\overline{F_n}$, the probability $q_{I,n}^M$ that image $I$ contains condition $c_n$ (simply noted $q_{I,n}$ in the absence of ambiguity) is defined as follows (see Fig.~\ref{fig:detailedPipeline}):
\begin{equation}
  q_{I,n} = \frac{F_n(\tau_I)}{F_n(\tau_I) + \overline{F_n}(\tau_I)} \;\;\;.
  \label{eq:probabilityFunction}
\end{equation}
A strong similarity can be noted between Equation (\ref{eq:tSNE}) of t-SNE and Equations (\ref{eq:probabilityDensityFunctions}) and (\ref{eq:probabilityFunction}) of probability function estimation: the main difference is the change of emphasis from sample-level in t-SNE to class-level in probability function estimation.

\subsection{Detecting Rare Conditions in one Image}
\label{sec:RareConditionDetection}

One challenge arises once we need to process a new image: Equation (\ref{eq:probabilityFunction}) is only theoretical. Indeed, the $\mathcal{S}' \rightarrow \mathcal{S}''$ projection based on t-SNE cannot be written in closed form. It is only defined for the development samples (i.e. $\forall I \in \mathcal{D}_R$), but it does not allow projection of new samples in the output $\mathcal{S}''$ feature space. This limitation does not apply to the projection from the image space to space $\mathcal{S}$ (CNN) or from space $\mathcal{S}$ to space $\mathcal{S}'$ (PCA).

In order to bypass this lack of expression, the following pipeline is proposed to determine the probability that condition $c_n$ is present in a new image $I\in\mathcal{D}_T$ (see Fig.~\ref{fig:detailedPipeline}):
\begin{enumerate}
  \item $I$ is processed by the CNN and the output $\gamma_I$ of the penultimate layer are computed (see Sections \ref{sec:DeepLearningFrequentConditionDetection} and \ref{sec:FeatureSpaceDefinition}).
	\item The PCA-based $\mathcal{S} \rightarrow \mathcal{S}'$ projection is applied to obtain $\pi_I$ (see Section \ref{sec:FeatureSpaceDimensionReduction}).
	\item A $K$-nearest neighbor regression is performed to approximate $q_{I,n}$. The search for the $K$ nearest neighbors $(V_k)_{k=1..K}$ is performed in $\mathcal{S}'$. The reference samples are the $(\langle \pi_J, q_{J,n} \rangle)_{J \in \mathcal{D}_R}$ couples, where the $q_{J,n}$ values are computed exactly through Equation (\ref{eq:probabilityFunction}). The approximate prediction $\hat{q}_{I,n}$ is given by:
\end{enumerate}
\begin{equation}
  \hat{q}_{I,n} = \frac{1}{ \displaystyle\sum_{k=1}^K{ \frac{1}{\|\pi_I-\pi_{V_k}\|}} } \sum_{k=1}^K{\frac{q_{V_k,n}}{\|\pi_I-\pi_{V_k}\|}} \;\;\;.
  \label{eq:probabilityKNN}
\end{equation}
The $\hat{q}_{I,n}$ prediction is a weighted arithmetic mean of exact predictions $q_{V_k,n}$, computed for the neighbors $V_k$ of $I$ in $\mathcal{S}'$: the weight assigned to $V_k$ is inversely proportional to the distance between $I$ and $V_k$ in $\mathcal{S}'$.

Note that the above pipeline for test images is differentiable, provided that the $K$ nearest neighbors of $I$ are considered constant. It can thus be implemented as a differentiable processing graph $G$ (see Fig.~\ref{fig:detailedPipeline}), stacking the following operations:
\begin{enumerate}
  \item the CNN up to the penultimate layer,
  \item the PCA-based $\mathcal{S} \rightarrow \mathcal{S}'$ linear projection,
  \item and the regression of Equation (\ref{eq:probabilityKNN}).
\end{enumerate}
This property allows heatmap generation (see Section \ref{sec:Visualization}) and may also allow fine-tuning of CNN weights.

In summary, the probability that a condition $c_n$ is present in any image $I$ can be estimated using Equation (\ref{eq:probabilityKNN}). If $n \leq M$, two probabilities of presence can be used: either $\hat{q}_{I,n}^M$ or $p_{I,n}^M$ (see Section \ref{sec:DeepLearningFrequentConditionDetection}).

\section{Experiments in the OPHDIAT Dataset}
\label{sec:Experiments}

We have presented a probabilistic framework for detecting rare conditions in images. This framework is now applied to DR screening in images from the OPHDIAT network.

\subsection{The OPHDIAT Dataset}
\label{sec:OphdiatDataset}

\begin{figure*}[!t]
  \begin{center}
    \includegraphics[height=.9\textwidth]{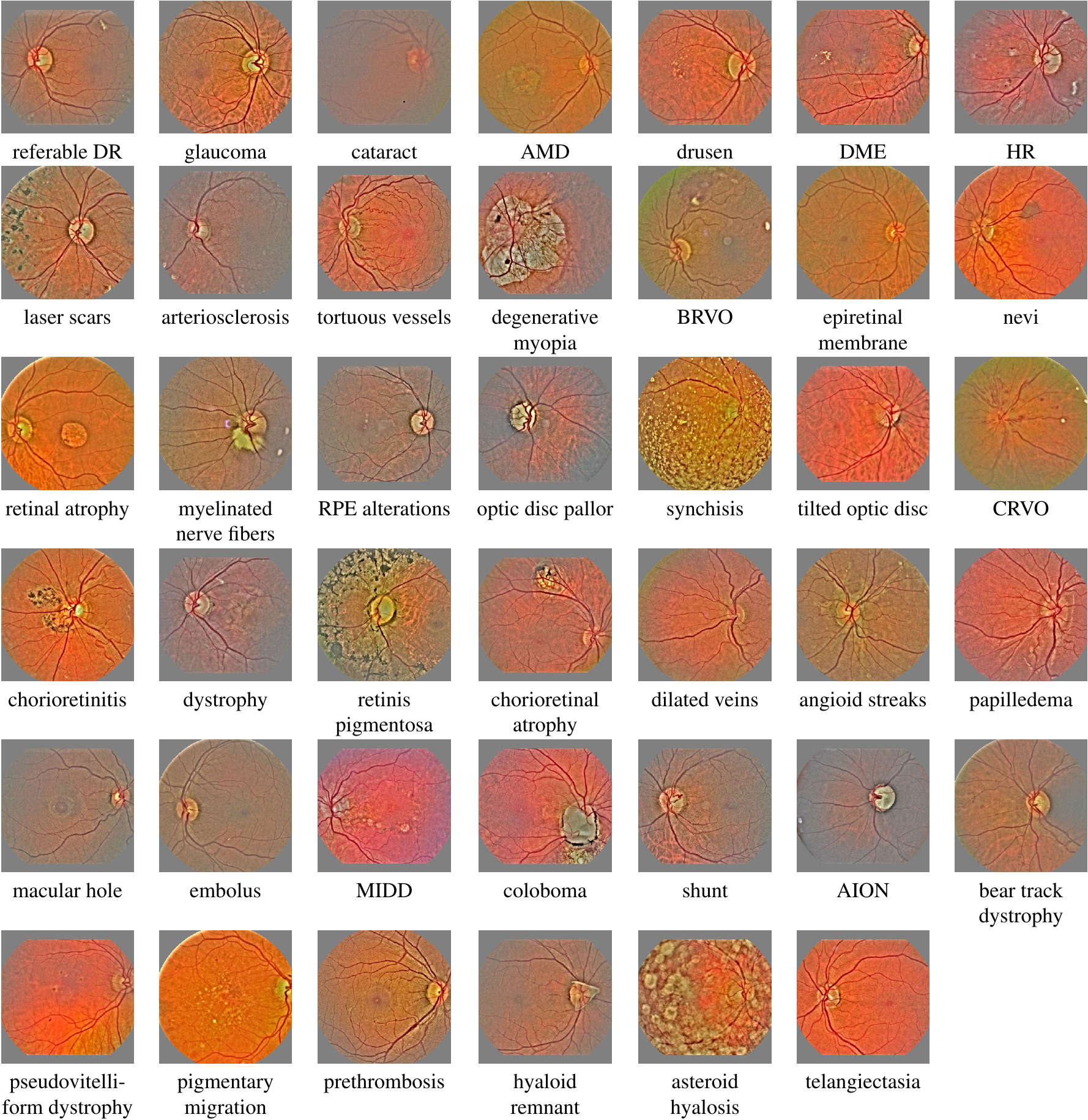}
  \end{center}
  \caption{Examples of images from each targeted condition. For improved visualization, the preprocessed images (see Section \ref{sec:ImagePreprocessing}) are reported. DR: diabetic retinopathy; AMD: age-related macular degeneration; DME: diabetic macular edema; HR: hypertensive retinopathy; BRVO: branch retinal vein occlusion; RPE: retinal pigment epithelium; CRVO: central retinal vein occlusion; MIDD: maternally inherited diabetes and deafness; AION: anterior ischemic optic neuropathy.}
  \label{fig:Conditions}
\end{figure*}

The OPHDIAT network consists of 40 screening centers located in 22 diabetic wards of hospitals, 15 primary health-care centers and 3 prisons in the Ile-de-France area \citep{massin_ophdiat:_2008}. Each center is equipped with one of the following 45$^\circ$ digital non-mydriatic cameras: Canon CR-DGI or CR2 (Tokyo, Japan), Topcon TRC-NW6 or TR-NW400 (Rotterdam, The Netherlands). Two photographs were taken per eye, one centered on the posterior pole and the other on the optic disc, and transmitted to the central server for interpretation and storage. From 2004 to the end of 2017, a total of 164,660 screening procedures were performed and 763,848 images were collected.

\subsection{Ground Truth Annotations}
\label{sec:Conditions}

Each screening exam was analyzed by one of the seven certified ophthalmologists of the OPHDIAT Reading Center, through a web interface, in order to generate a structured report \citep{massin_ophdiat:_2008}. This structured report includes the grading of diabetic retinopathy (DR) in each eye. It also indicates the presence or suspicion of presence of a few other pathologies in each eye. In addition to the structured report, the ophthalmologist also indicated his or her findings in free-form text. For the purpose of this study, these reports were analyzed by a retina specialist and 41 conditions were identified ($N=41$ --- see Fig. \ref{fig:Conditions}). Ground truth annotations were obtained for each eye by combining structured information and manually-extracted textual information. Next, annotations were assigned to images thanks to our laterality identification algorithm \citep{QuellecInstantautomaticdiagnosis2019}: this algorithm assigns a label, `left eye' or `right eye', to each image using an ensemble of CNNs. One limitation of this approach is that ophthalmologists may not have written all their findings. To ensure that ``normal images'' are indeed non-pathological, normal images were visually inspected and images containing anomalies were discarded: a total of 16,955 normal images, out of 18,000 inspected images, were included. A total of 115,159 images were included in dataset $\mathcal{D}$.

\subsection{Concordance of Annotations}
\label{sec:ConcordanceAnnotations}

To ensure the efficacy and safety of the OPHDIAT program, quality insurance procedures were set up \citep{massin_ophdiat:_2008}. Before starting the grading task, all screeners underwent a training program provided by two senior ophthalmologists. Then, interpretive accuracy was verified on a monthly basis: every month, 5\% of the photographs were selected and automatically merged with new patient data for a double interpretation. In terms of DR diagnosis, interpretations were concordant for 96.85\% of the photographs on average \citep{QuellecInstantautomaticdiagnosis2019}. In case of disagreement, one of the senior ophthalmologists makes the final interpretations.

Concordance of annotations was not evaluated for all forty other conditions individually: some conditions are so rare that it would require double interpretation on a very large subset of the OPHDIAT dataset. Instead, we evaluated concordance of normal versus pathological retina identification on a subset of images from 5,000 patients. Those images were selected randomly among patients without DR, according to the OPHDIAT reader. Those images were interpreted again by another retina specialist. Annotations were concordant for 4,509 patients (90.18\%).

\subsection{Image Preprocessing}
\label{sec:ImagePreprocessing}

\begin{figure}[!t]
  \begin{center}
    \includegraphics[width=\textwidth]{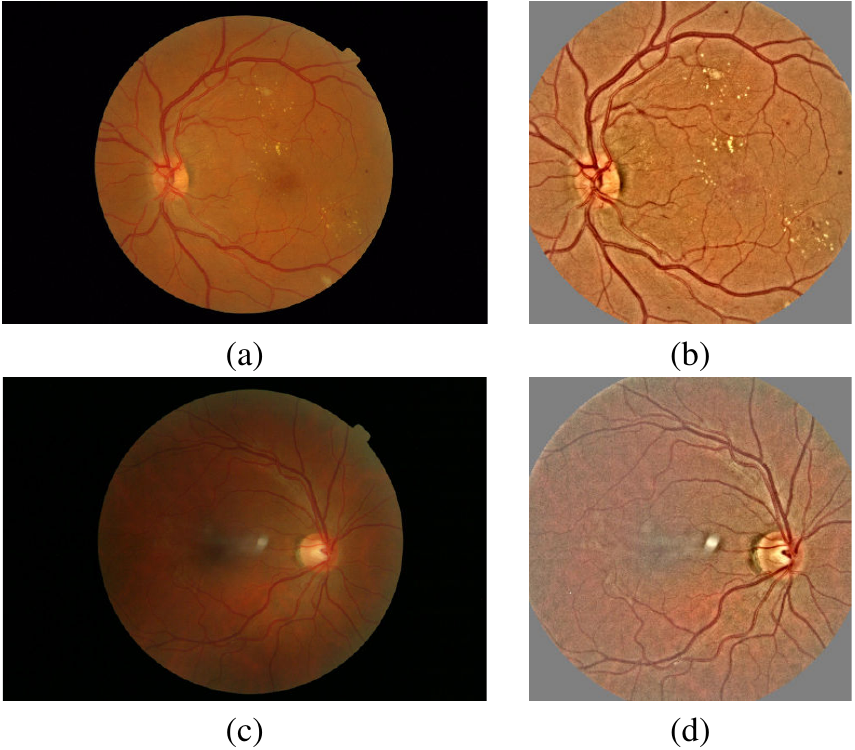}
  \end{center}
  \caption{Fundus photograph preprocessing. Original images (a) and (c) are transformed into (b) and (d).}
  \label{fig:FundusPhotographPreprocessing}
\end{figure}

\begin{table}
  \caption{Average classification scores (AUC) in the test subset for frequent conditions ($n \leq M_0=11$) using various CNN architectures. Scores for single CNNs are in the diagonal; scores for CNN pairs are above. Architectures selected in the validation subset are in bold.}
  \begin{center}
    \begin{tabular}{c|c@{\hskip 0.1in}c@{\hskip 0.1in}c@{\hskip 0.1in}c@{\hskip 0.1in}c@{\hskip 0.1in}c}
                  & \RotText{Inception-3} & \RotText{Inception-4} & \RotText{ResNet-50} & \RotText{ResNet-101} & \RotText{ResNet-152} & \RotText{NASNet-A} \\
      \hline
      Inception-3 & \textbf{0.966} & \textbf{0.963} & 0.935 & 0.924 & 0.920 & 0.921 \\
      Inception-4 & & \textbf{0.960} & 0.930 & 0.926 & 0.931 & 0.927 \\
      ResNet-50   & & & 0.912 & 0.924 & 0.925 & 0.916 \\
      ResNet-101  & & & & 0.916 & 0.919 & 0.918 \\
      ResNet-152  & & & & & 0.890 & 0.895 \\
      NASNet-A    & & & & & & 0.882 \\
    \end{tabular}
  \end{center}
  \label{tab:classificationScoresFrequent}
\end{table}

Because fundus photographs were acquired with various cameras, the size and appearance of images were normalized to allow device-independent analysis. For size normalization, a square region of interest was defined around the camera's field of view; this region of interest was then resized to 299$\times$ 299 pixels. For appearance normalization, illumination variations throughout and across images were attenuated. This step was performed in the YCrCb color space. In this color space, components Cr and Cb contain chrominance information: these components were left unchanged. Component Y represents luminance: this channel was normalized to compensate for illumination variations. For that purpose, the image background was estimated using a Gaussian kernel with a large kernel size (standard deviation: 5 pixels). Next, this background image was subtracted from the Y channel. Finally, the obtained image was converted to an RGB image.

We used a similar model for diabetic retinopathy (DR) screening, except that each channel in the RGB color space was normalized independently, as described above for the Y channel \citep{QuellecDeepimagemining2017a}. Although suitable for DR screening, that representation proved inefficient to detect pigmentation conditions in particular.

\subsection{Performance Assessment}

Because a patient may suffer from several conditions simultaneously, a variable number of conditions may be visible in each image: zero, one or more (up to 41). So the problem we are addressing is a multilabel classification problem. This is equivalent to 41 independent binary classification problems: one per condition. For each condition $c_n$, classification performance was defined as the area under the ROC (Receiver-Operating Characteristic) curve, noted AUC: the binary label used for a given image $I$ is $y_{I,n}$ (see Section \ref{sec:Notations}).

\subsection{Learning, Validation and Testing}
\label{sec:LearningValidationTesting}

In the presence of numerous and highly unbalanced conditions, dividing the dataset into learning, validation and test subsets is critical. The following strategy was proposed to 1) distribute the data between subsets, 2) train and validate the models and 3) test the selected models; one model is selected per condition based on validation scores. The proposed data distribution strategy, described hereafter, ensures that two images from the same patient were assigned to the same subset (either $\mathcal{D}_L$, $\mathcal{D}_V$ or $\mathcal{D}_T$).

For the purpose of training CNNs, a ``balanced'' dataset $\mathcal{B}_M$ was created in such a way that, ideally, all frequent conditions were equally represented. For each condition $c \in \{c_M, c_{M-1}, ..., c_2, c_1\}$, images with condition $c$ were selected at random until all images containing $c$ had been selected or until the number of selected images with condition $c$ reached 1,500. Images containing rare conditions were excluded from this selection. 5,000 normal images were also selected at random. The size of these balanced datasets ranges from 17,205 images (for $\mathcal{B}_{11}$) to 21,973 images (for $\mathcal{B}_{41}$).

The learning subset $\mathcal{D}_L$ was populated by $80\%$ of $\mathcal{B}_M$. The validation subset $\mathcal{D}_V$ was populated by $10\%$ of $\mathcal{B}_M$ plus $20\%$ of $\mathcal{D} \setminus \mathcal{B}_M$. The test subset $\mathcal{D}_T$ was populated by $10\%$ of $\mathcal{B}_M$ plus $80\%$ of $\mathcal{D} \setminus \mathcal{B}_M$. Taking validation and test images from $\mathcal{B}_M$ ensures that all conditions $c_n$, $n \leq M$, can be validated and tested. Both $\mathcal{B}_M$ and $\mathcal{D} \setminus \mathcal{B}_M$ were split at random, while ensuring that the distribution of conditions is similar in each fold. As an illustration, the distribution of conditions in each subset is given in Table \ref{tab:SubsetSize} for $M=17$.

\begin{table}
  \caption{Frequency of each condition in the full dataset ($\mathcal{D}$), in the ``balanced'' dataset ($\mathcal{B}_M$), and in the learning, validation and testing subsets ($\mathcal{D}_L$, $\mathcal{D}_V$ and $\mathcal{D}_T$, respectively), when $M=17$ conditions are considered frequent.}
  \begin{center}
    \begin{tabular}{l@{\hskip 0.05in}|c@{\hskip 0.05in}|c@{\hskip 0.05in}|c@{\hskip 0.05in}c@{\hskip 0.05in}c}
    condition ($c_n$)           & $\mathcal{D}$ & $\mathcal{B}_M$ & $\mathcal{D}_L$ & $\mathcal{D}_V$ & $\mathcal{D}_T$ \\
    \hline
    normal images               & 16955 & 5000 & 3992 & 2890  & 10073 \\
    referable DR	              & 65560 & 3036 & 2406 & 12819 & 50335 \\
    glaucoma                    & 10624 & 1500 & 1176 & 1992  & 7456 \\
    cataract                    & 3540  & 1502 & 1253 & 546   & 1741 \\
    AMD                         & 3173  & 1637 & 1336 & 442   & 1395 \\
    drusen                      & 3164  & 1502 & 1179 & 511   & 1474 \\
    DME                         & 3024  & 1471 & 1168 & 463   & 1393 \\
    HR                          & 3018  & 1502 & 1184 & 453   & 1381 \\
    laser scars                 & 1450  & 1018 & 796  & 204   & 450 \\
    arteriosclerosis            & 1235  & 802  & 629  & 161   & 445 \\
    tortuous vessels            & 1222  & 1060 & 830  & 161   & 231 \\
    degenerative myopia         & 1209  & 1086 & 859  & 145   & 205 \\
    BRVO                        & 752   & 532  & 429  & 117   & 206 \\
    epiretinal membrane         & 674   & 557  & 447  & 78    & 149 \\
    nevi                        & 628   & 557  & 466  & 55    & 107 \\
    retinal atrophy	            & 546   & 433  & 340  & 67    & 139 \\
    myelinated nerve fibers	    & 531   & 472  & 372  & 62    & 97 \\
    RPE alterations	            & 508   & 446  & 370  & 50    & 88 \\
    \hline
    optic disc pallor	          & 455   & 0    & 0    & 91    & 364 \\
    synchisis	                  & 366   & 0    & 0    & 73    & 293 \\
    tilted optic disc	          & 334   & 0    & 0    & 67    & 267 \\
    CRVO                        & 297   & 0    & 0    & 59    & 238 \\
    chorioretinitis	            & 294   & 0    & 0    & 59    & 235 \\
    dystrophy	                  & 217   & 0    & 0    & 43    & 174 \\
    retinis pigmentosa          & 183   & 0    & 0    & 37    & 146 \\
    chorioretinal atrophy	      & 182   & 0    & 0    & 36    & 146 \\
    dilated veins	              & 165   & 0    & 0    & 33    & 132 \\
    angioid streaks	            & 145   & 0    & 0    & 29    & 116 \\
    papilledema	                & 99    & 0    & 0    & 20    & 79 \\
    macular hole	              & 78    & 0    & 0    & 16    & 62 \\
    embolus	                    & 74    & 0    & 0    & 15    & 59 \\
    MIDD	                      & 70    & 0    & 0    & 14    & 56 \\
    coloboma	                  & 52    & 0    & 0    & 11    & 41 \\
    shunt	                      & 51    & 0    & 0    & 10    & 41 \\
    AION	                      & 43    & 0    & 0    & 9     & 34 \\
    bear track dystrophy	      & 42    & 0    & 0    & 8     & 34 \\
    pseudovitelliform          	& \multirow{2}{*}{29} & \multirow{2}{*}{0} &
                                  \multirow{2}{*}{0} & \multirow{2}{*}{6} & \multirow{2}{*}{23} \\
    dystrophy                   &       &      &      &       &  \\
    pigmentary migration        & 28    & 0    & 0    & 6     & 22 \\
    prethrombosis	              & 28    & 0    & 0    & 5     & 23 \\
    hyaloid remnant	            & 16    & 0    & 0    & 4     & 12 \\
    asteroid hyalosis	          & 15    & 0    & 0    & 3     & 12 \\
    telangiectasia	            & 15    & 0    & 0    & 3     & 12 \\
    \end{tabular}
  \end{center}
  \label{tab:SubsetSize}
\end{table}

The usual validation and testing strategy was followed for strongly-supervised CNN-based detectors (providing the $p_{I,n}$ predictions of section \ref{sec:DeepLearningFrequentConditionDetection}). However, validating and testing the detectors (providing the $\hat{q}_{I,n}$ predictions of section \ref{sec:RareConditionDetection}) is more challenging: ideally, we would need a fourth independent subset $\mathcal{D}_R$ of reference images. In order to maximize the size of the testing, validation and reference subsets, which is particularly critical for rare conditions, a different solution was used instead. First, validation relied on a 10-fold cross-validation strategy: for each fold, probability functions were built using $90\%$ of $\mathcal{D}_V$ as reference images ($\mathcal{D}_R$), and $\hat{q}_{I,n}$ predictions were computed for the remaining $10\%$. Similarly, a 10-fold cross-testing strategy was followed: for each fold, probability functions were built using $\mathcal{D}_V$ plus $90\%$ of $\mathcal{D}_T$ as reference images, and $\hat{q}_{I,n}$ predictions were computed for the remaining $10\%$ of $\mathcal{D}_T$. In both cross-validation and cross-testing, images from the same patient were all assigned to the same fold, to avoid evaluation biases. Finally, in both cross-validation and cross-testing, a single ROC curve was built for $c_n$ by joining all the $\hat{q}_{I,n}$ predictions, computed in all ten folds, together with the associated $y_{I,n}$ target labels.

\subsection{Parameter Selection}
\label{sec:CNNArchitectureSelection}

The choice of $M$ frequent conditions is arbitrary. In initial experiments, we set $M=M_0=11$: $M_0$ was chosen such that $f_n \geq 1000$, $\forall n \leq M_0$. The following CNN architectures were investigated: Inception-v3 \citep{SzegedyRethinkinginceptionarchitecture2016}, Inception-v4 \citep{SzegedyInceptionv4InceptionResNetimpact2017}, ResNet-50, ResNet-101 and ResNet-152 \citep{HeDeepresiduallearning2016}, and NASNet-A \citep{ZophLearningtransferablearchitectures2018}. These CNNs were pre-trained on the public ImageNet dataset and fine-tuned on the $\mathcal{D}_L$ learning subset. The TF-slim image classification library was used.\footnote{\url{https://github.com/tensorflow/models/tree/master/research/slim}} The combination of two CNNs was also investigated: in that case, their penultimate layers were concatenated to define the initial $\mathcal{S}$ feature space. An experiment involving the $M_0=11$ most frequent conditions was performed to select the most promising architectures. Average AUC scores for the $M_0$ most frequent conditions on the $\mathcal{D}_V$ validation subset are reported in Table \ref{tab:classificationScoresFrequent} for each architecture. This experiment reveals that three architectures lead to particularly good classification performance: Inception-v3, Inception-v4 and ``Inception-v3 + Inception-v4''. We only considered those three architectures in subsequent experiments.

\begin{figure*}[!t]
  \begin{center}
    \begin{tabular}{c@{\hskip 0.095in}c@{\hskip 0.095in}c@{\hskip 0.095in}c}
      \includegraphics[height=0.212\textwidth]{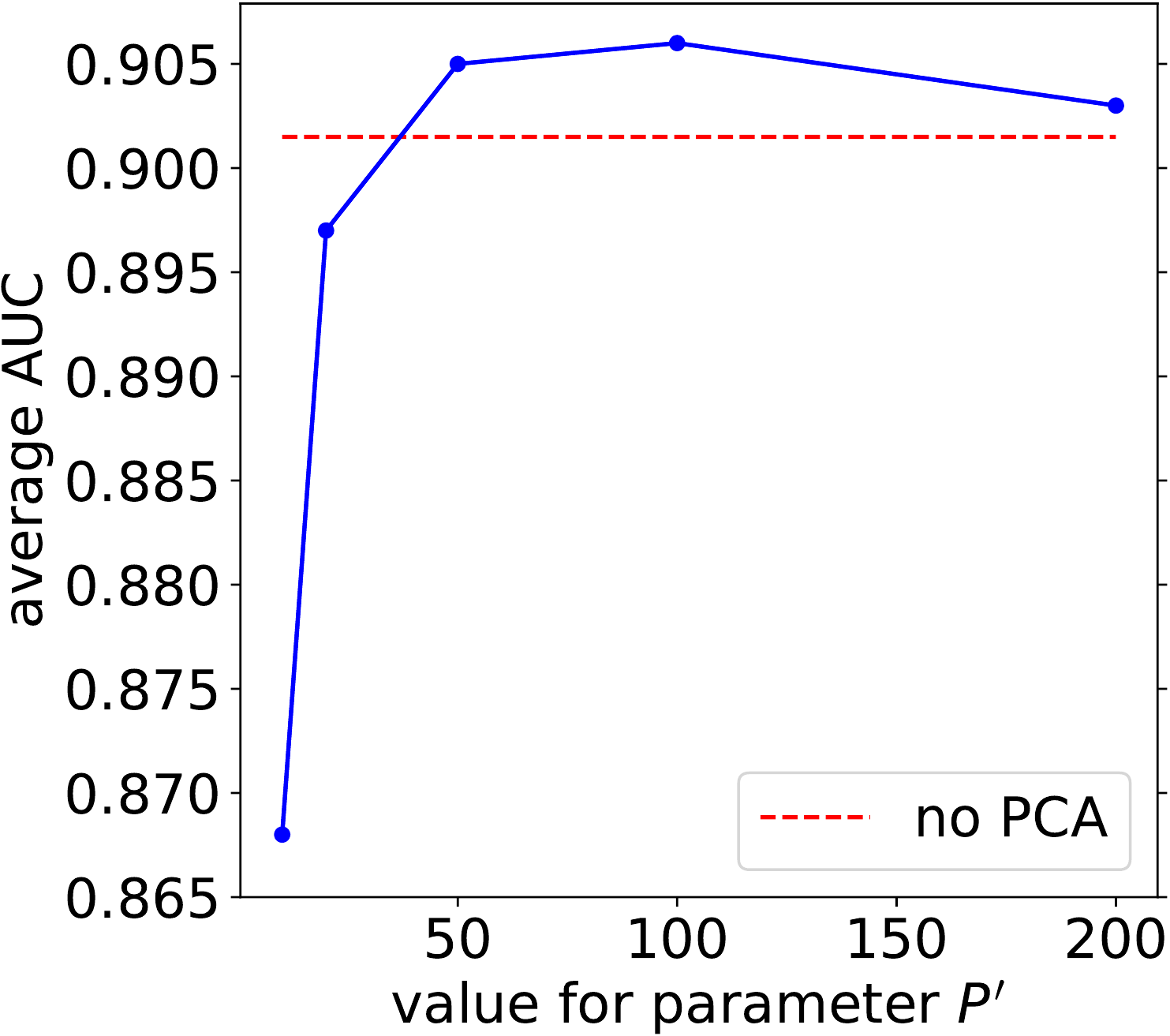} &
      \includegraphics[height=0.212\textwidth]{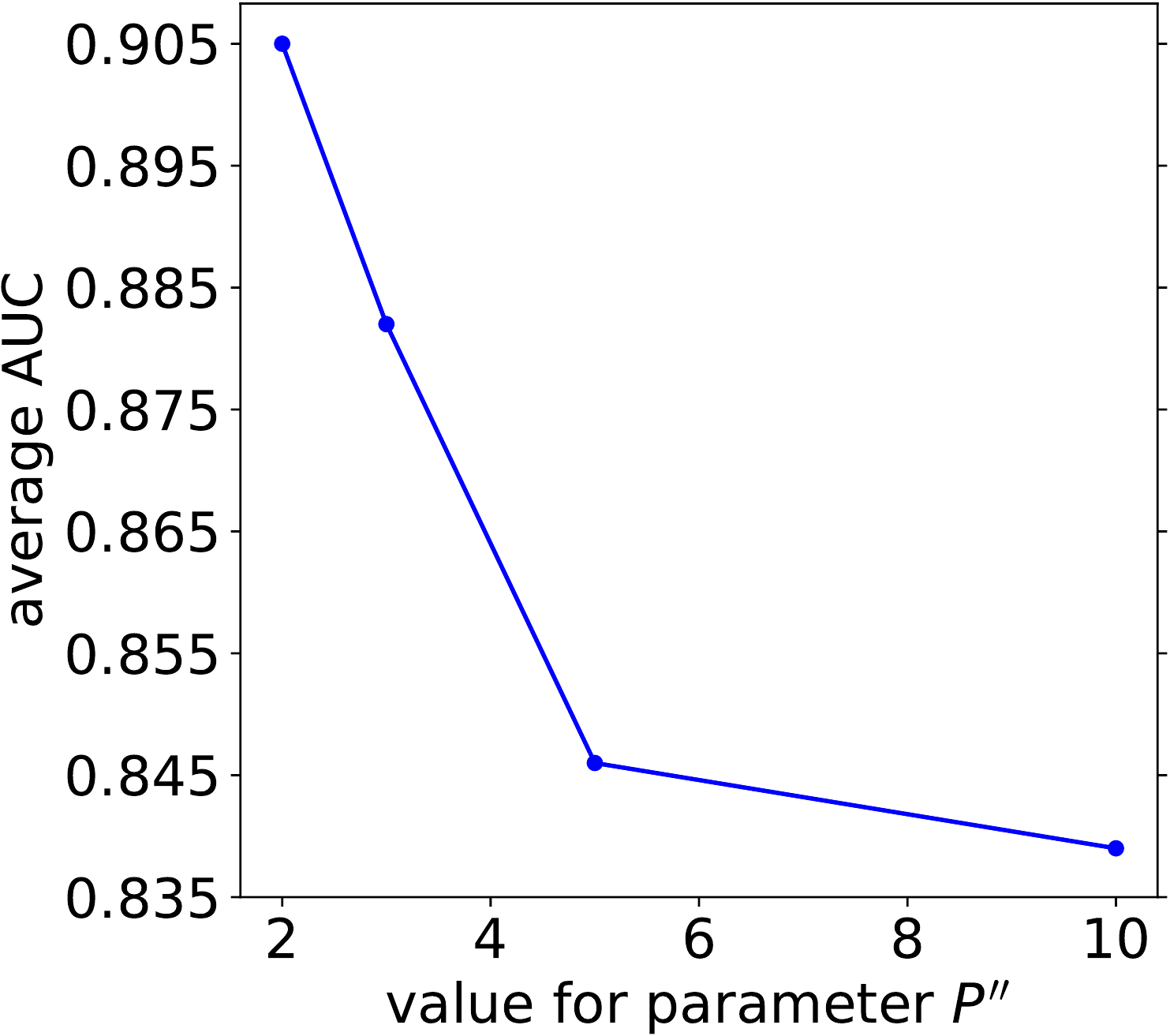} &
      \includegraphics[height=0.212\textwidth]{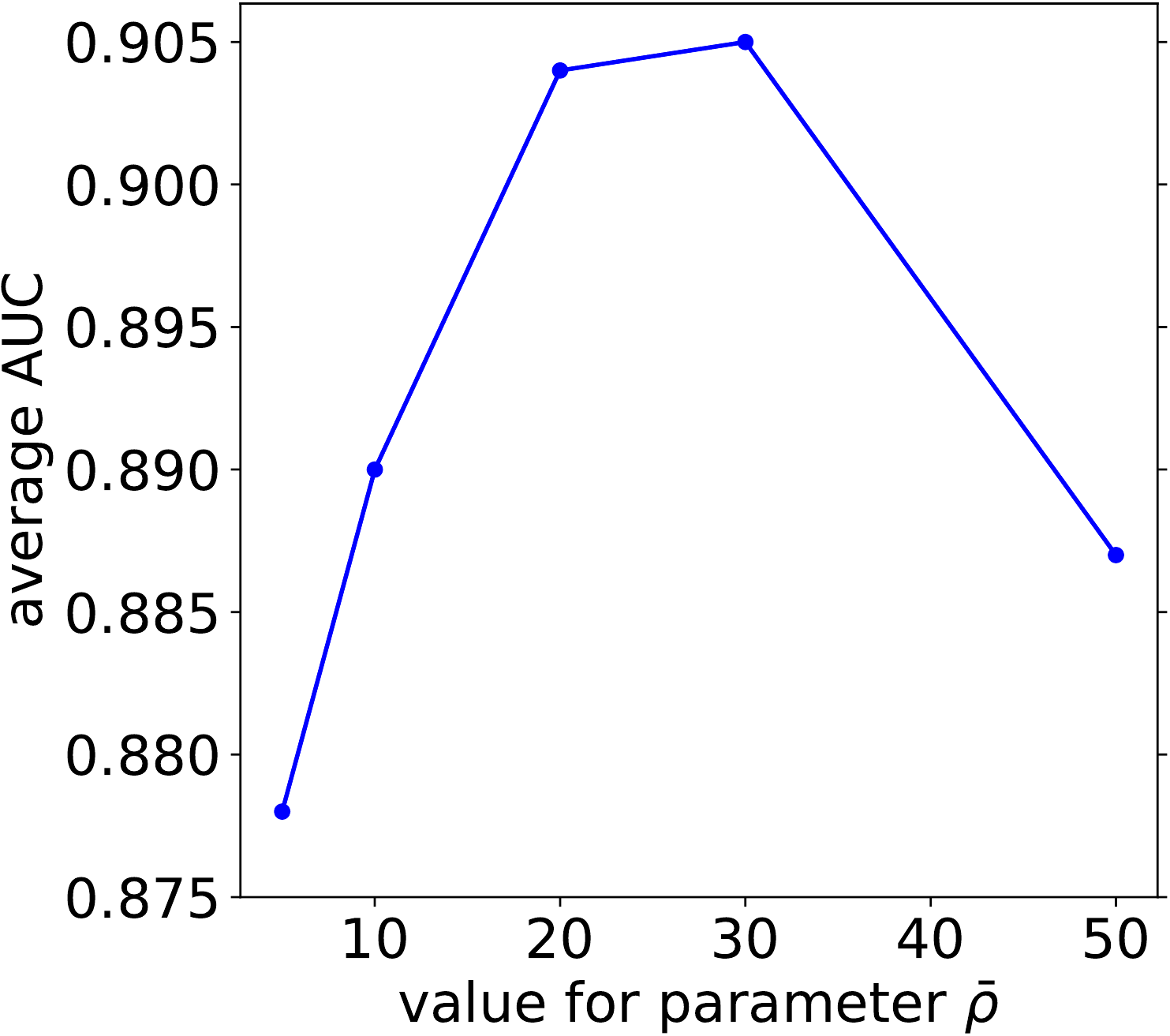} &
      \includegraphics[height=0.212\textwidth]{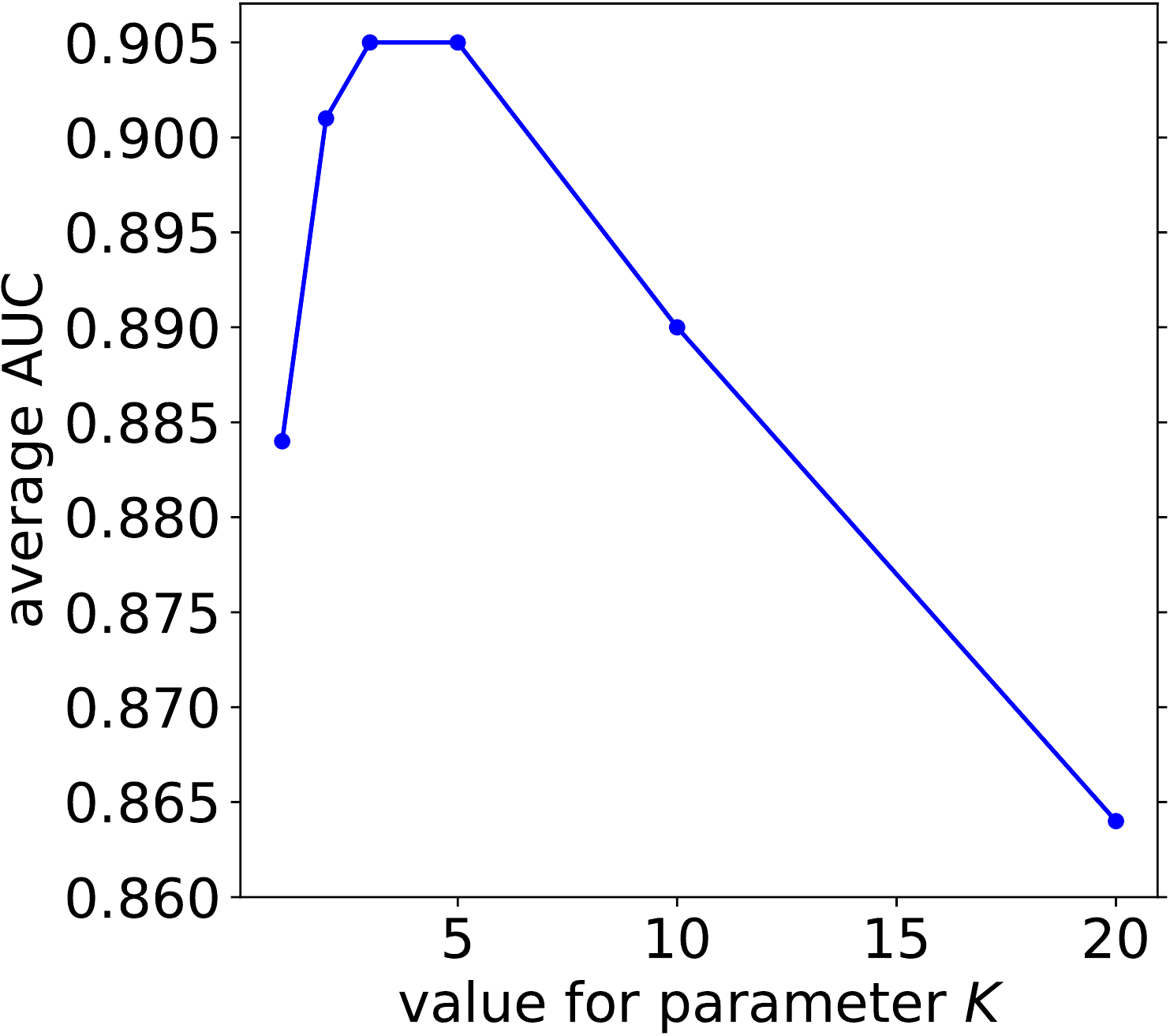}
    \end{tabular}
  \end{center}
  \caption{Classification performance on the test subset for one CNN as a function of the main algorithm parameters: a) the dimension $P'$ of the feature space generated by PCA, b) the dimension $P''$ of the feature space generated by t-SNE, c) perplexity $\bar{\rho}$ in t-SNE and d) the number $K$ of neighbors used to computed the $\hat{q}_{I,n}$ predictions. Each subfigure reports the average AUC on the test subset, as a function of the investigated parameter. The non-investigated parameters are set to the selected value ($P' = 50$, $P'' = 2$, $\bar{\rho} = 30$, $K = 3$). We also evaluated a scenario where no PCA preprocessing is performed ($\sim P'=\infty$): in that case, k-NN regression is performed in $\mathcal{S}$ during inference on test images (see Fig. \ref{fig:detailedPipeline}); an average AUC of 0.9015 was obtained. The CNN used in all these experiments is Inception-v3, trained when $M = 17$ conditions are considered frequent (see Fig. \ref{fig:DensityFunctions}).}
  \label{fig:impactParameters}
\end{figure*}

\begin{figure*}[!t]
  \begin{center}
    \includegraphics[width=\textwidth]{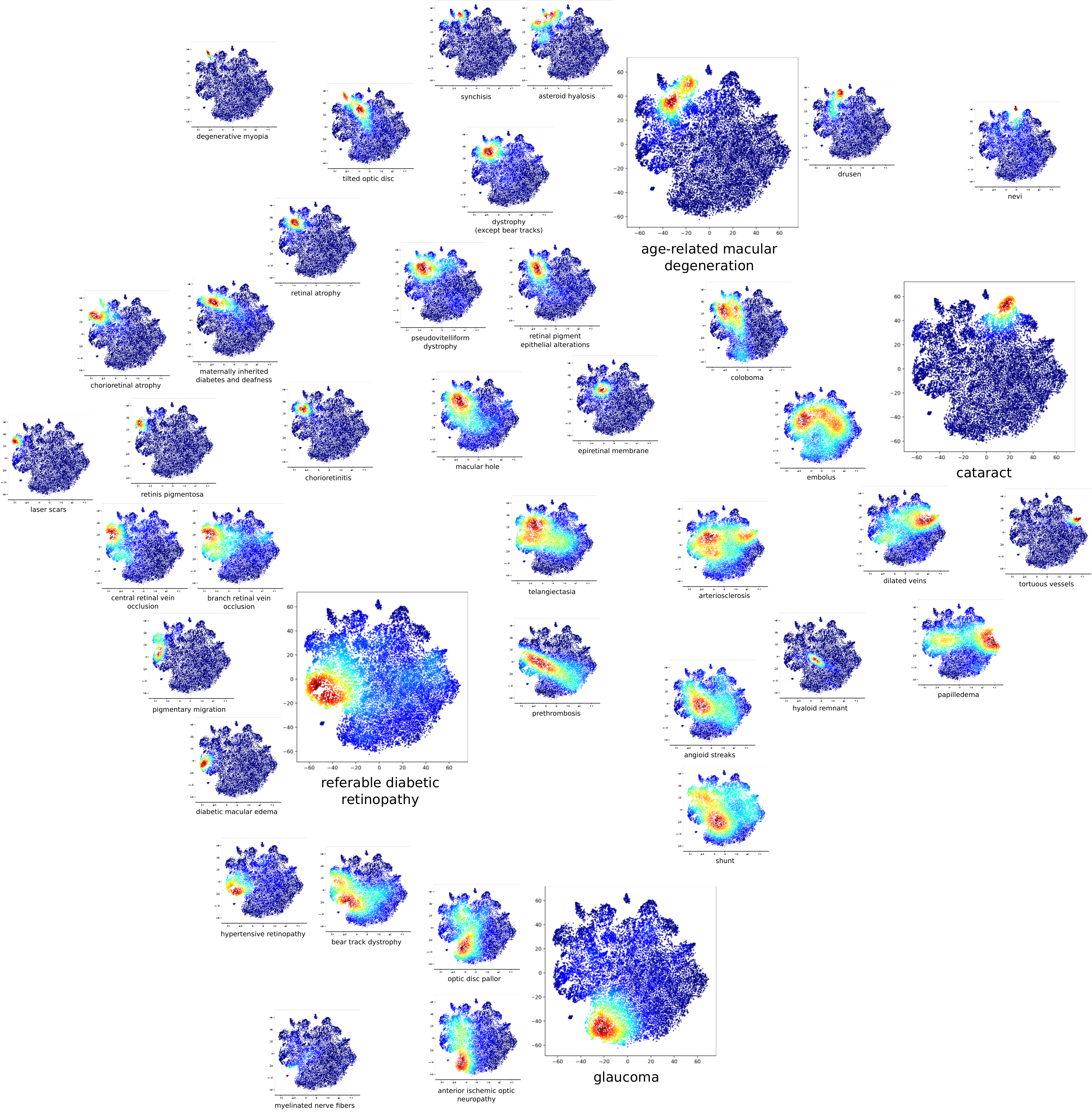}
  \end{center}
  \caption{Probability density functions obtained using Inception-v3 when $M = 17$ conditions are considered frequent. One subfigure is obtained per condition: for each location in feature space, red (respectively blue) indicates a large (respectively low) probability density for that condition. Figures associated with the four most frequent conditions are emphasized. A widespread probability density function (e.g. `embolus') indicates that images with or without the condition could not be separated well. Conversely, a narrow distribution (e.g. `degenerative myopia') indicates good separation. Additionally, two overlapping distributions (e.g. `synchisis' and `asteroid hyalosis' --- two types of vitreous opacities) indicates that the associated conditions could not be separated well; for convenience, subfigures are grouped by similarity of the probability density functions. It can be noted that `drusen' and `pseudovitelliform dystrophy' overlap with `age-related macular degeneration (AMD)', which makes sense since these conditions are generally associated with AMD.}
  \label{fig:DensityFunctions}
\end{figure*}

Two important parameters also had to be set:
\begin{itemize}
  \item the dimension $P''$ of the reduced feature space generated by t-SNE (for visualization, $P''$ is generally set to 2 or 3, but higher values can be used),
  \item the number $K$ of neighbors to approximate the $\hat{q}_{I,n}$ predictions in Equation (\ref{eq:probabilityKNN}).
\end{itemize}
These parameters were chosen to maximize classification performance on the validation subset using $M=M_0$. The optimal parameter values were: $P''=2$ and $K=3$. Other dimension reduction parameters were set to commonly used values: the number of dimensions after PCA was set to $P' = 50$ (see Section \ref{sec:FeatureSpaceDimensionReduction}) and perplexity in t-SNE was set to $\bar{\rho}=30$ (see Section \ref{sec:tSNE}).\footnote{\url{https://scikit-learn.org/stable/modules/generated/sklearn.manifold.TSNE.html}} The influence of these four parameters on classification performance is illustrated in Fig. \ref{fig:impactParameters} for one CNN. As an illustration, the probability density functions obtained for that CNN, with the selected parameter values, are shown in Fig. \ref{fig:DensityFunctions}.

In subsequent experiments, multiple values for $M$ were investigated. We varied $M$ from $M_0=11$ to $N=41$ by steps of 6 conditions: $M \in \{11,17,23,29,35,41\}$. A different value for $M$ and a different CNN architecture were selected for each condition: the combinations maximizing the AUC on the $\mathcal{D}_V$ validation subset were selected.

\subsection{Detection Performance}

\begin{figure*}[!t]
  \begin{center}
    \begin{tabular}{ccc}
    \includegraphics[height=0.3\textwidth]{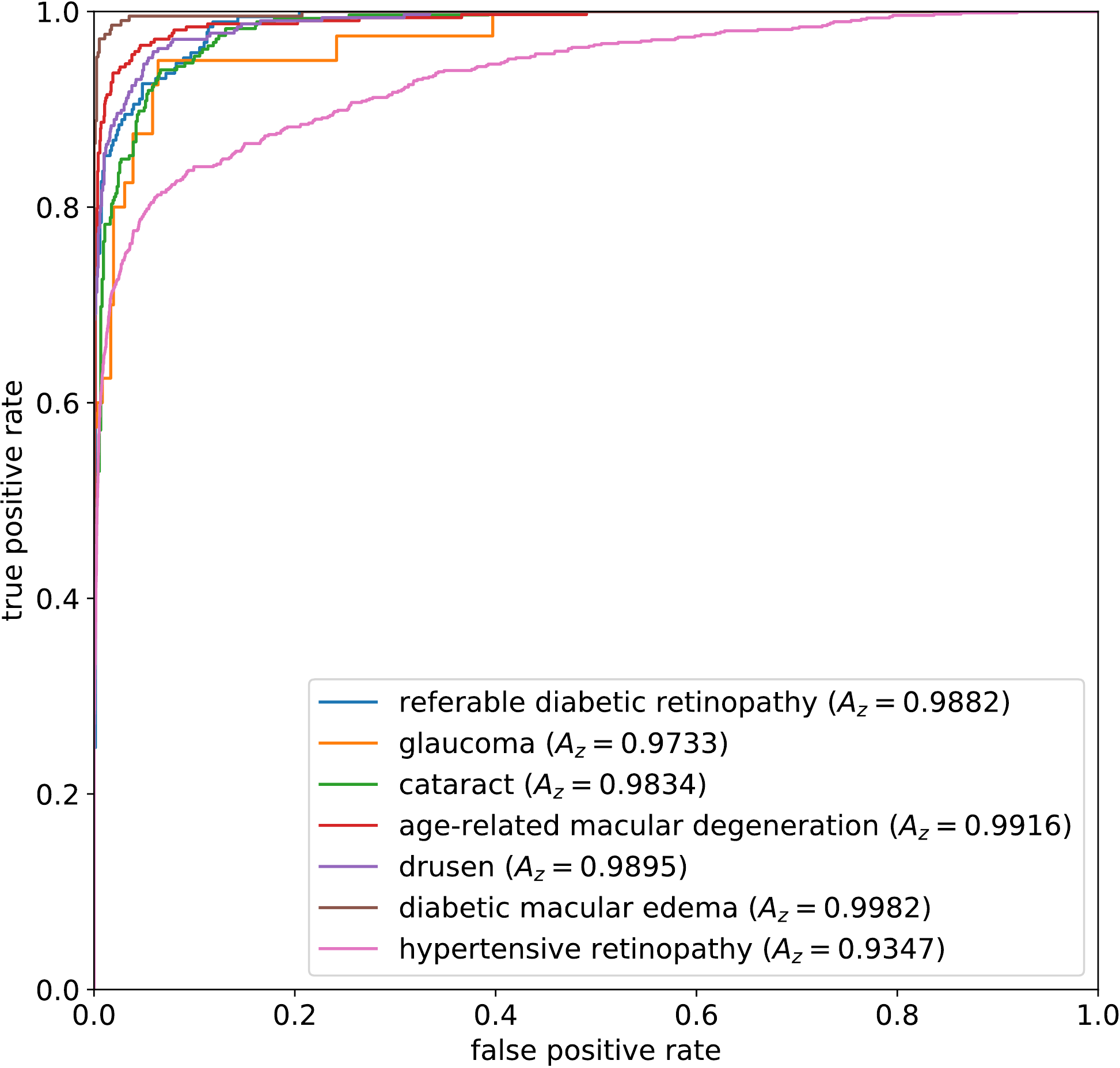} &
    \includegraphics[height=0.3\textwidth]{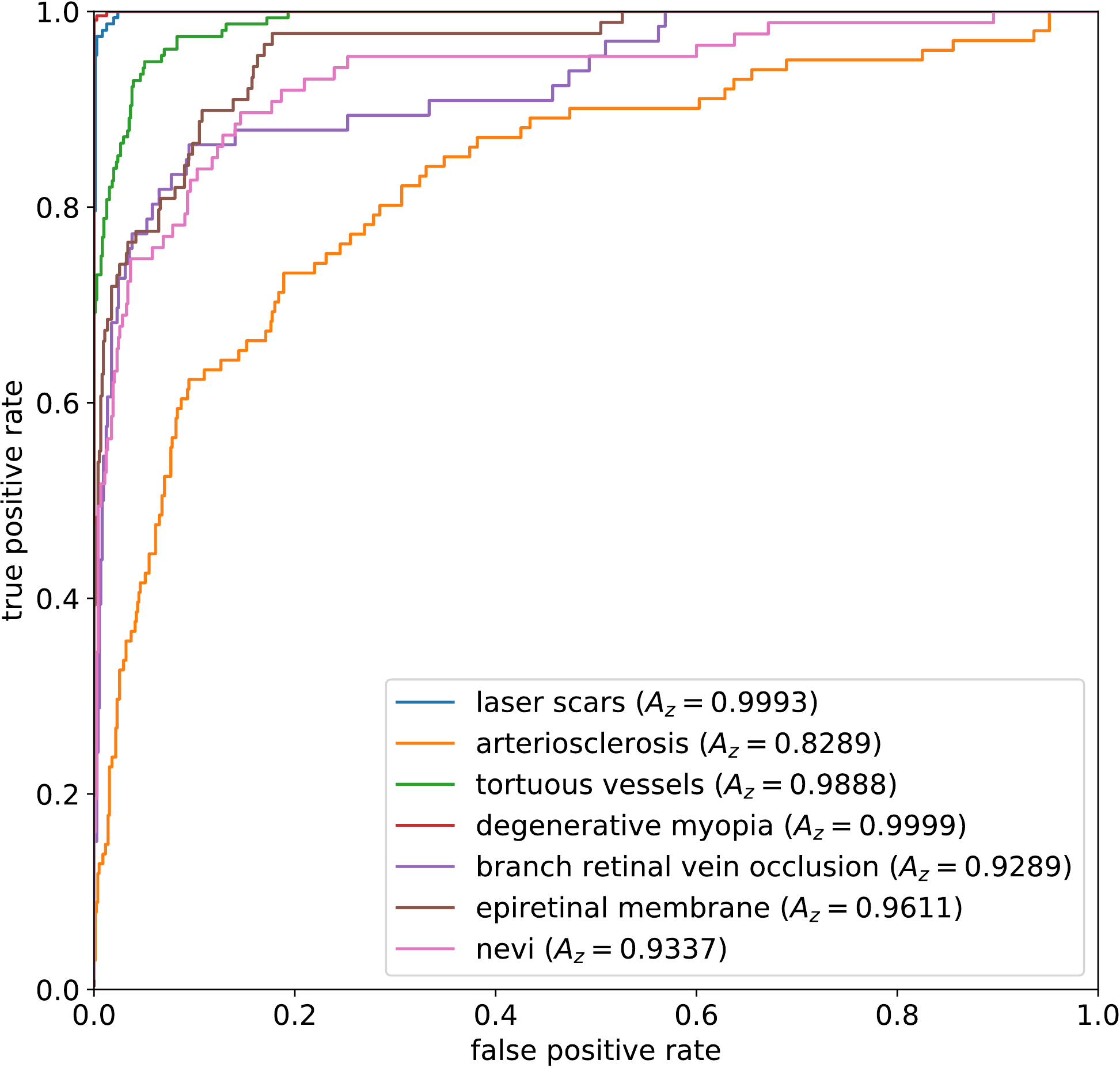} &
    \includegraphics[height=0.3\textwidth]{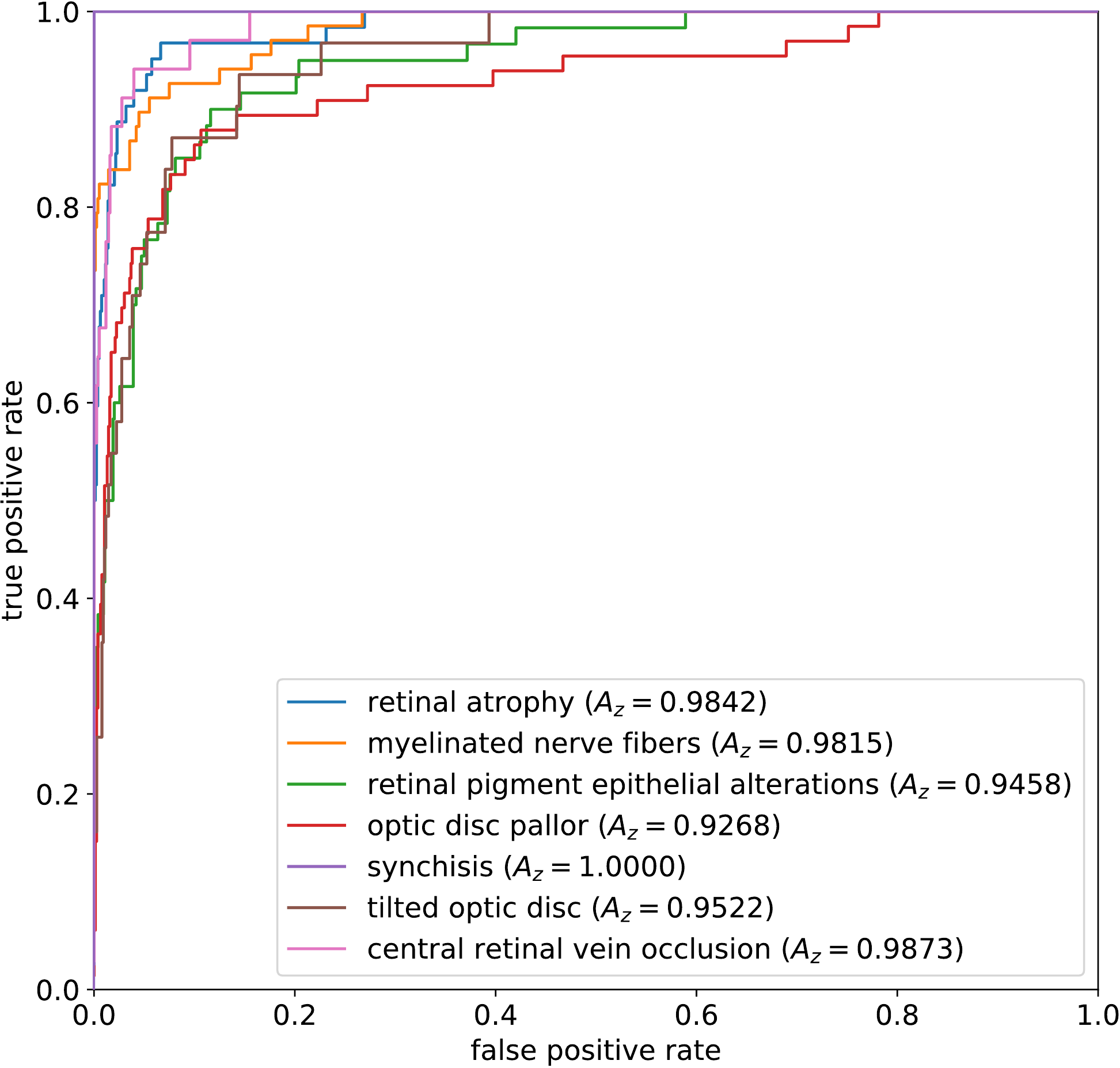} \\
    \includegraphics[height=0.3\textwidth]{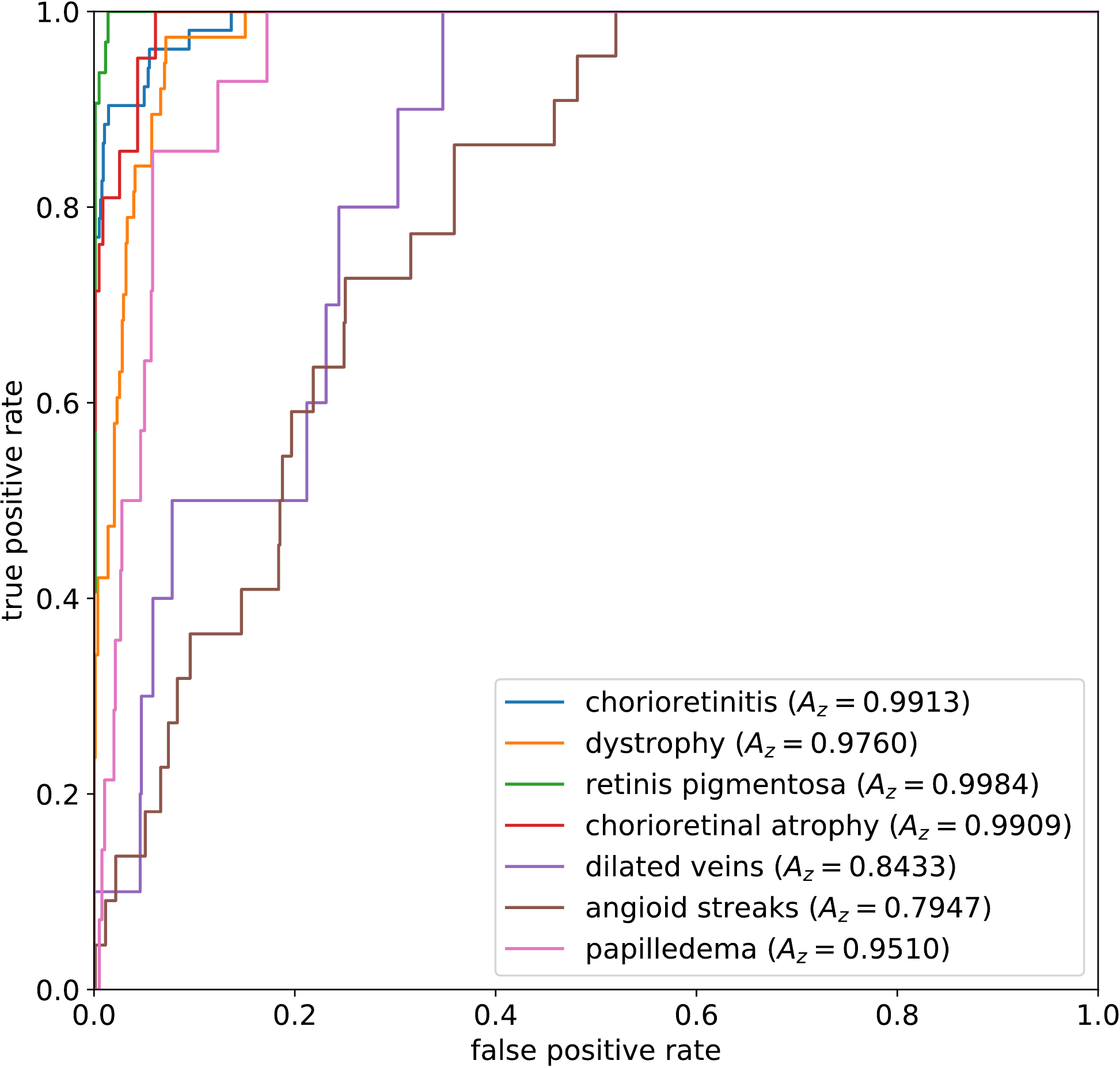} &
    \includegraphics[height=0.3\textwidth]{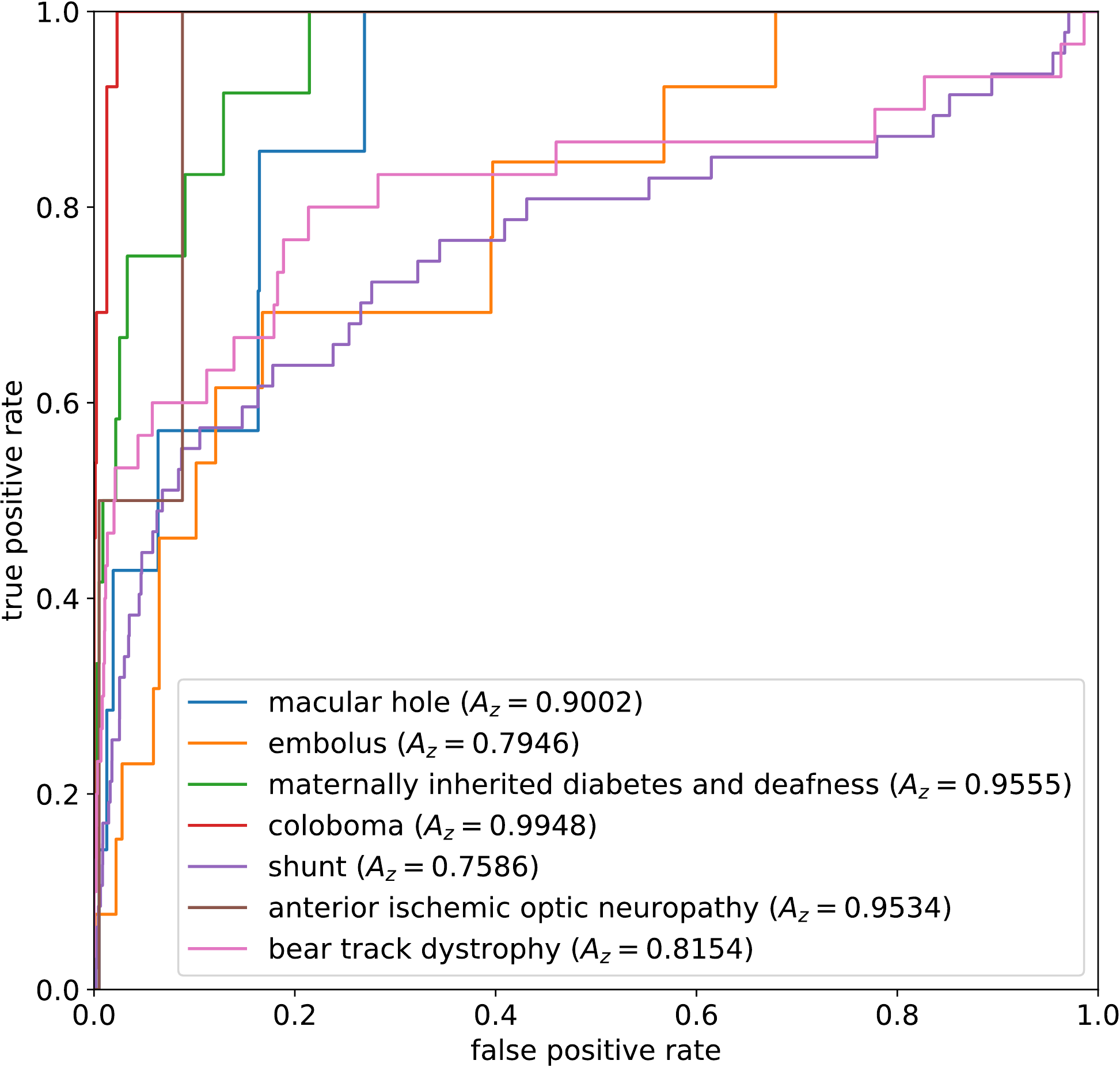} &
    \includegraphics[height=0.3\textwidth]{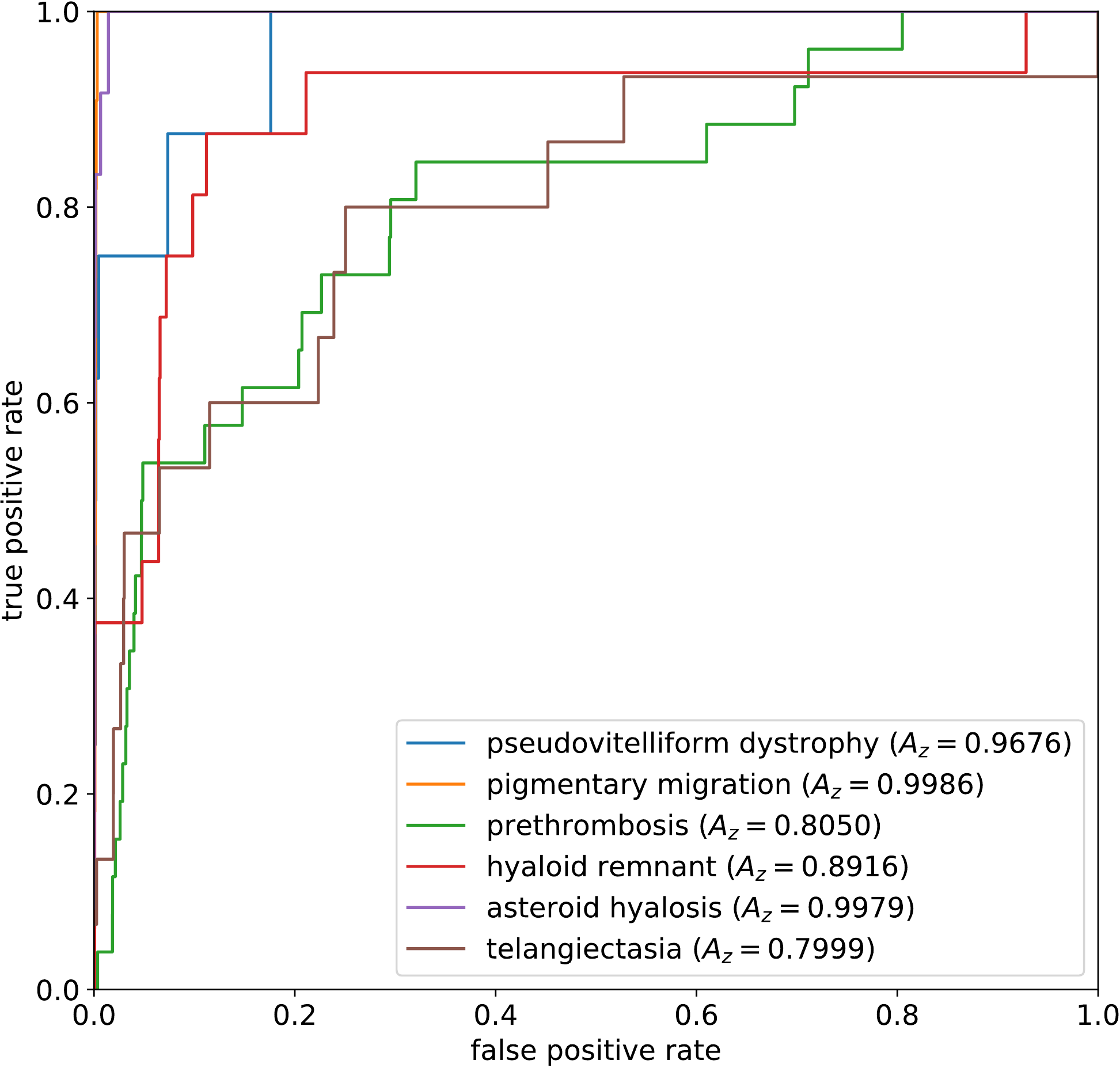} \\
    \end{tabular}
  \end{center}
  \caption{ROC curves on the test subset for each condition.}
  \label{fig:rocCurves}
\end{figure*}

\begin{figure*}[!t]
  \begin{center}
    \includegraphics[width=.8\textwidth]{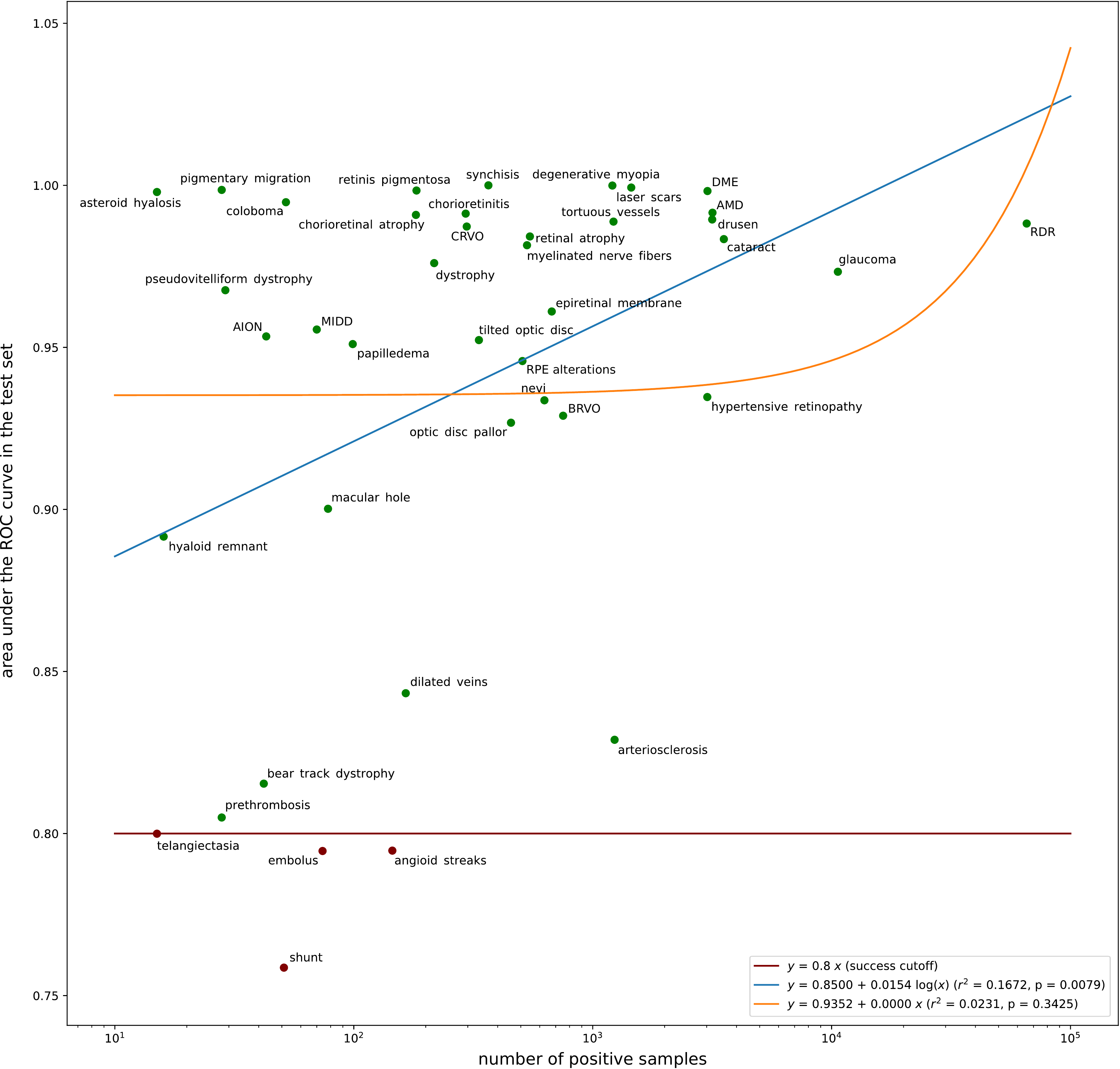}
  \end{center}
  \caption{Detection performance of conditions $c_n$ on the test subset as a function of frequency $f_n$. An area under the ROC curve less than 0.8 is considered a failure.}
  \label{fig:AucFrequency}
\end{figure*}

The ROC analysis of the proposed solution on the test subset is reported in Fig. \ref{fig:rocCurves}. The influence of condition frequency on the area under the ROC curve (AUC) is illustrated in Fig. \ref{fig:AucFrequency}. In both figures, a different model is used for each condition: the one maximizing the AUC on the validation subset. We observe that detection performance is poorly correlated with the frequency of a condition.

Because a different model was trained for each condition, computation times can be an issue. Therefore, we studied how fast a test image can be processed. Computation times were measured on a desktop computer with one Intel Xeon E5-1650 v2 Hexa-core 3.50 GHz CPU and one Nvidia GeForce GTX 1070 GPU. On average, running the 41 models on a single image takes 52 seconds without the GPU and 5.3 seconds using the GPU.

\subsection{Heatmap Generation}
\label{sec:Visualization}

\begin{figure*}[!t]
  \begin{center}
    \includegraphics[width=0.9\textwidth]{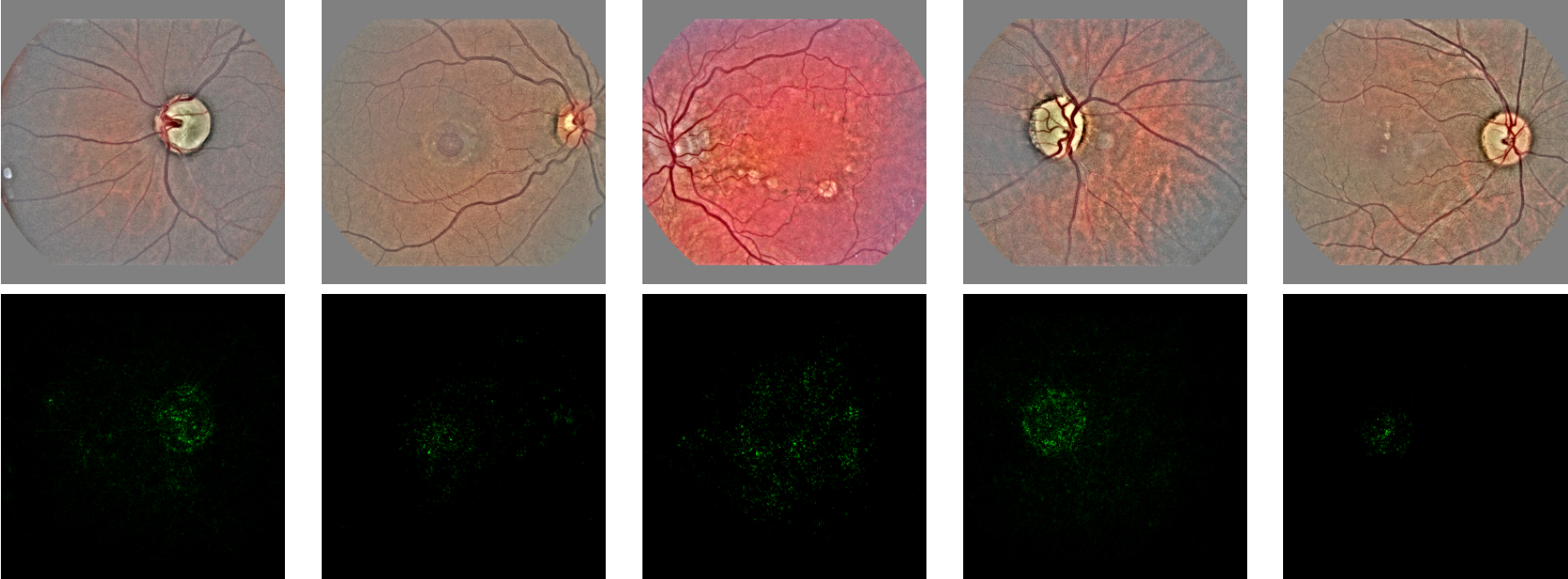}
  \end{center}
  \caption{Heatmap generation. Examples of images are given in the first row. From left to right, these images present an anterior ischemic optic neuropathy, a macular hole, maternally inherited diabetes and deafness, optic disc pallor and retinal pigment epithelium (RPE) alterations. Heatmaps are given in the second row for those conditions. Black means zero; positive values are in green. The CNN of Fig. \ref{fig:DensityFunctions} was used (Inception-v3 CNN --- $M = 17$): with the exception of RPE alterations (last column), those conditions are thus unknown to this CNN.}
  \label{fig:Heatmaps}
\end{figure*}

To understand what a detector looks for in a given image $I$, we can measure how much each pixel $I_{xy}$ contributes to its prediction: the result is a heatmap the size of $I$ showing the detected structures. A solution was proposed in the case where the predictions rely on $p_{I,n}^M$ \citep{QuellecDeepimagemining2017a}. That solution extends sensitivity analysis \citep{simonyan_deep_2014}: the idea is to compute the gradient of the model predictions with respect to each input pixel, using the backpropagation algorithm. When predictions rely on $\hat{q}_{I,n}^M$, the backpropagation algorithm can also be used: it is applied to the differentiable processing graph $G$ of Section \ref{sec:RareConditionDetection}. The contribution $\xi_{xyc}$ of each pixel $I_{xy}$, for condition $c$, is determined as follows:
\begin{equation}
  \xi_{xyc} = \left| \frac{\partial G(m \circ I, c)}{\partial m_{xy}} \right| \;\;\;,
  \label{eq:heatmapDensity}
\end{equation}
where $m$ denotes a matrix the size of $I$ filled with ones, and $\circ$ denotes the Hadamard product. Heatmap examples for conditions unknown to the CNNs (i.e. for rare conditions) are given in Fig. \ref{fig:Heatmaps}.

\subsection{Comparison with Other Machine Learning Frameworks}
\label{sec:ComparisonOtherMachineLearningFrameworks}

The proposed framework is now compared with competing ML frameworks, namely Siamese networks \citep{KochSiameseneuralnetworks2015}, a popular few-shot learning algorithm, transfer learning and multitask learning. The same CNN architectures (Inception-v3, Inception-v4 and ``Inception-v3 + Inception-v4'') were considered in all experiments.

In the reference few-shot learning algorithm, the similarity between two images $I$ and $J$ was defined using Siamese networks \citep{KochSiameseneuralnetworks2015}. For a fair comparison, the basis network inside the Siamese networks was replaced by one of the three selected CNN architectures. The outputs of the penultimate CNN layer, i.e. the $\gamma_I$ and $\gamma_J$ vectors, were used to compute the similarity between $I$ and $J$. This similarity is defined as a logistic regression of the absolute difference between $\gamma_I$ and $\gamma_J$ \citep{KochSiameseneuralnetworks2015}. For training the Siamese networks, $I$ and $J$ were considered to match if at least one condition was present in both images. To detect $c_n$ in a test image, the average similarity to validation images containing condition $c_n$ was used: it proved more efficient than considering the maximal similarity \citep{KochSiameseneuralnetworks2015}.

For transfer learning, CNNs were trained for the $M_0=11$ most frequent conditions. Then, these CNNs were fine-tuned to detect each of the remaining 30 conditions individually. For multitask learning, CNNs were trained to detect the 41 conditions altogether.

Results are reported in Table \ref{tab:ComparisonFrameworks}. Statistical comparisons between frameworks were performed using repeated measures ANOVA, using the MedCalc Statistical Software version 19.0.6 (MedCalc Software bvba, Ostend, Belgium). Results are reported in Table \ref{tab:anova}.

\begin{table}
  \caption{Comparison between ML frameworks in terms of AUC on the test subset. Abbreviations are listed in the legend of Fig. \ref{fig:Conditions}. The best AUC for each condition is in bold.}
  \begin{center}
    \begin{tabular}{l@{\hskip 0.075in}|c@{\hskip 0.075in}c@{\hskip 0.075in}c@{\hskip 0.075in}c}
    condition ($c_n$)           & \RotText{proposed} & \RotText{Siamese networks} & \RotText{transfer learning} & \RotText{multitask learning} \\
    \hline
    referable DR	              & \textbf{0.9882} & 0.7422 & \textbf{0.9882} & 0.9251 \\
    glaucoma                    & \textbf{0.9733} & 0.7567 & \textbf{0.9733} & 0.9600 \\
    cataract                    & \textbf{0.9834} & 0.7947 & 0.9780 & 0.9754 \\
    AMD                         & \textbf{0.9916} & 0.7441 & \textbf{0.9916} & 0.9717 \\
    drusen                      & \textbf{0.9895} & 0.8304 & \textbf{0.9895} & 0.9470 \\
    DME                         & \textbf{0.9982} & 0.8709 & 0.9959 & 0.9973 \\
    HR                          & \textbf{0.9347} & 0.8087 & 0.9299 & 0.8568 \\
    laser scars                 & \textbf{0.9993} & 0.7647 & \textbf{0.9993} & 0.9971 \\
    arteriosclerosis            & \textbf{0.8289} & 0.8110 & 0.7998 & 0.8083 \\
    tortuous vessels            & \textbf{0.9888} & 0.7603 & \textbf{0.9888} & 0.9774 \\
    degenerative myopia         & \textbf{0.9999} & 0.8726 & \textbf{0.9999} & 0.9973 \\
    \hline
    BRVO                        & \textbf{0.9289} & 0.8285 & 0.8020 & 0.8613 \\
    epiretinal membrane         & \textbf{0.9611} & 0.8521 & 0.9162 & 0.9456 \\
    nevi                        & \textbf{0.9337} & 0.8250 & 0.8035 & 0.6311 \\
    retinal atrophy	            & \textbf{0.9842} & 0.7605 & 0.8293 & 0.9178 \\
    myelinated nerve fibers	    & \textbf{0.9815} & 0.8255 & 0.9462 & 0.9059 \\
    RPE alterations	            & \textbf{0.9458} & 0.8600 & 0.9177 & 0.8818 \\
    optic disc pallor	          & \textbf{0.9268} & 0.8537 & 0.8794 & 0.8878 \\
    synchisis	                  & \textbf{1,0000} & 0.9091 & 0.9932 & 0.9991 \\
    tilted optic disc	          & \textbf{0.9522} & 0.8568 & 0.9031 & 0.9258 \\
    CRVO                        & 0.9873 & 0.8325 & 0.9028 & \textbf{0.9879} \\
    chorioretinitis	            & \textbf{0.9913} & 0.7635 & 0.9555 & 0.9862 \\
    dystrophy	                  & \textbf{0.9760} & 0.9493 & 0.8303 & 0.9058 \\
    retinis pigmentosa          & \textbf{0.9984} & 0.9889 & 0.9740 & 0.9967 \\
    chorioretinal atrophy	      & \textbf{0.9909} & 0.8435 & 0.8405 & 0.9714 \\
    dilated veins	              & 0.8433 & \textbf{0.8804} & 0.8093 & 0.8063 \\
    angioid streaks	            & 0.7947 & \textbf{0.9314} & 0.7594 & 0.8090 \\
    papilledema	                & \textbf{0.9510} & 0.9403 & 0.9363 & 0.9192 \\
    macular hole	              & \textbf{0.9002} & 0.8784 & 0.6734 & 0.7404 \\
    embolus	                    & 0.7946 & \textbf{0.8565} & 0.6690 & 0.6916 \\
    MIDD	                      & \textbf{0.9555} & 0.9132 & 0.9206 & 0.9270 \\
    coloboma	                  & \textbf{0.9948} & 0.7188 & 0.9346 & 0.6446 \\
    shunt	                      & 0.7586 & \textbf{0.8380} & 0.6818 & 0.6782 \\
    AION	                      & \textbf{0.9534} & 0.9330 & 0.9108 & 0.8330 \\
    bear track dystrophy	      & \textbf{0.8154} & 0.6921 & 0.6245 & 0.5912 \\
    pseudovitelliform          	& \multirow{2}{*}{\textbf{0.9676}} & \multirow{2}{*}{0.9412} & \multirow{2}{*}{0.9176} & \multirow{2}{*}{0.9157} \\
    dystrophy                   &                 &        &        &        \\
    pigmentary migration        &	\textbf{0.9986} & 0.7750 & 0.8714 & 0.9251 \\
    prethrombosis	              & \textbf{0.8050} & 0.6688 & 0.5123 & 0.4325 \\
    hyaloid remnant	            & \textbf{0.8916} & 0.7247 & 0.8000 & 0.6426 \\
    asteroid hyalosis	          & \textbf{0.9979} & 0.7635 & 0.9327 & 0.8382 \\
    telangiectasia	            & \textbf{0.7999} & 0.6423 & 0.3982 & 0.4236 \\
    \hline
    \hline
    average ($\forall n$)       & \textbf{0.9380} & 0.8245 & 0.8654 & 0.8545 \\
    average ($\forall n > 11$)  & 0.9260 & 0.8349 & 0.8282 & 0.8207 \\
    \end{tabular}
  \end{center}
  \label{tab:ComparisonFrameworks}
\end{table}

\begin{table}
  \caption{Comparison between ML frameworks using repeated measures ANOVA. Comparisons consider either all conditions or rare conditions ($n > M_0 = 11$). Significant differences are in bold.}
  \begin{center}
    \begin{tabular}{l@{\hskip 0.075in}|l@{\hskip 0.075in}|c@{\hskip 0.075in}c@{\hskip 0.075in}c}
                                & & \RotText{Siamese networks} & \RotText{transfer learning} & \RotText{multitask learning} \\
    \hline
    \multirow{3}{*}{$\forall n$}
    & proposed          & \textbf{$<$0.0001} & \textbf{$<$0.0001} & \textbf{0.0001} \\
    & Siamese networks  &                    & 0.3507             & 0.9401 \\
    & transfer learning &                    &                    & 1.0000 \\
    \hline
    \multirow{3}{*}{$\forall n > 11$}
    & proposed          & \textbf{0.0002} & \textbf{$<$0.0001} & \textbf{0.0002} \\
    & Siamese networks  &                 & 1.0000             & 1.0000 \\
    & transfer learning &                 &                    & 1.0000 \\
    \hline
    \end{tabular}
  \end{center}
  \label{tab:anova}
\end{table}

\section{Discussion and Conclusions}
\label{sec:DiscussionConclusions}

We have presented a new few-shot learning framework for detecting rare conditions in medical images using deep learning. This framework takes advantage of many annotations available for more frequent conditions in a large image dataset. This framework was successfully applied to the detection of 41 conditions in fundus photographs from the OPHDIAT diabetic retinopathy (DR) screening program.

This framework takes advantage of an interesting behavior of convolutional neural networks (CNNs): CNNs tend to cluster similar images in feature space, a phenomenon exploited in content-based image retrieval systems for instance \citep{ToliasParticularObjectRetrieval2016}. In our context, we observed that conditions unknown to the CNNs are also clustered in feature space (see Fig. \ref{fig:DensityFunctions}). A probabilistic framework, based on the t-SNE representation, was thus proposed to take advantage of this observation. Detection results are very good: an average area under the ROC curve (AUC) of 0.9380 was obtained (see Table \ref{tab:ComparisonFrameworks}). Detection performance is also good if we consider the rarest conditions alone: the average AUC only drops to 0.9260 when the 30 rarest pathologies are considered (see Table \ref{tab:ComparisonFrameworks}). More generally, we observed in Fig. \ref{fig:AucFrequency} that detection performance is poorly correlated with the frequency of a condition: $r^2 = 0.0231$ (or $r^2 = 0.1672$ using a logarithmic scale for frequency). Prior to the study, we established that a detector would be considered useful should the AUC exceed 0.8: that cutoff was reached for 37 conditions out of 41. In retrospect, eight conditions stand out as being particularly difficult to detect automatically: shunt, angioid streaks, embolus, telangiectasia, prethrombosis, bear track dystrophy, arteriosclerosis and dilated veins (see Fig. \ref{fig:AucFrequency}). Among these conditions, ophthalmologists consider that one is particularly difficult to detect in fundus photographs, namely telangiectasia, and three are associated with a poor reproducibility, namely embolus, arteriosclerosis and dilated veins. However, the other four are considered easy to detect. One reason for this poor automatic detection may be inadequate image preprocessing (too low image resolution after resizing, inadequate color normalization, etc.). Another explanation may be a too large difference with frequent conditions, which would explain that no discriminatory features are extracted by the CNNs.

In terms of computation times, the proposed framework is comparable with standard strongly-supervised CNNs: besides CNN inference, processing a test image simply involves a linear projection in PCA space ($\mathcal{S}'$) and a K-nearest neighbor search in that space. On average, detecting all conditions in one image without GPU takes 52 seconds. This is similar to computation times reported for commercial systems detecting diabetic retinopathy alone.\footnote{\url{www.eyediagnosis.co} --- \url{www.eyenuk.com}} Besides, this process can be parallelized easily in a cloud-based solution. Using a single CNN for all conditions is possible but not recommended, as performance is lower: for instance, AUC = 0.905 for Inception-v3 and $M = 17$ (see Fig. \ref{fig:impactParameters}). Training times are not impacted much either: it only takes a few additional minutes compared to CNN training alone. PCA helps reducing computation times during training and inference, as t-SNE (during training) and k-NN regression (during inference) are performed in a lower-dimensional space. It also increases performance (see Fig. \ref{fig:impactParameters}), possibly by addressing the curse of dimensionality in k-NN regression.

The proposed framework was compared to other candidate ML frameworks for detecting rare conditions: transfer learning, multitask learning and Siamese networks. First, it appears that the proposed approach significantly outperforms standard approaches like transfer learning and multitask learning (see Table \ref{tab:anova}). The comparison with transfer learning is particularly interesting: we show that, with similar complexity (one CNN model per condition), the proposed approach detects rare conditions ($n > 11$) significantly better. In transfer learning, we hypothesize that good properties learnt for the detection of frequent conditions are lost when fine-tuning for rare conditions. Worse, in multitask learning, the detection of frequent conditions, which is trained simultaneously, is negatively impacted. The comparison with Siamese networks is more contrasted: Siamese networks outperformed the proposed solution for four conditions. All these conditions are among the rarest ($n > M_0 = 11$) and three of them were poorly detected (AUC $<$ 0.8) by the proposed solution (see Table \ref{tab:ComparisonFrameworks}). Interestingly, the performance of Siamese networks \citep{KochSiameseneuralnetworks2015} proved to be highly independent of the frequency of a condition. However, Table \ref{tab:anova} shows that the proposed solution significantly outperforms Siamese networks, which have similar complexity (one CNN model per condition). In summary, the proposed solution clearly is the most relevant ML framework for the target task (see Table \ref{tab:anova}). However, Siamese networks, which also relies on similarity analysis in CNN feature space \citep{KochSiameseneuralnetworks2015}, is an interesting few-shot learning framework as well.

We believe the use of a visualization technique, namely t-SNE, for classification is an interesting feature of the proposed framework. In particular, we found that reducing the feature space to two dimensions, the value generally used for visualization, maximizes classification performance (see Section \ref{sec:CNNArchitectureSelection}). This can be explained in part by more reliable kernel density estimations in low-dimensional feature spaces \citep{ScottMultivariatedensityestimation1992}. One advantage is that we can conveniently browse the image dataset in a 2-D viewer and understand how the dataset is organized by CNNs. This could be used to show human readers similar images for decision support \citep{QuellecAutomatedassessmentdiabetic2011}. Although the probabilistic model is based on the t-SNE dimension reduction technique, which is expression-less, we designed it to be differentiable. This property allows heatmap generation, through sensitivity analysis \citep{QuellecDeepimagemining2017a}, for improved visualization. In particular, we can see that the pathological structures are well captured by the CNNs, even for conditions unknown to the CNNs (see Fig. \ref{fig:Heatmaps}).

The proposed framework is mostly unsupervised, which can be regarded as a limitation. We note, however, that it can easily be transformed into a supervised framework. The solution is to 1) approximate the t-SNE projection with a multilayer perceptron and 2) optimize this approximation to maximize the separation between probability density functions $F_n$ and $\overline{F_n}$ \citep{PatrickNonparametricfeatureselection1969}. The CNN weights can thus be optimized through the probabilistic model. However, the number of degrees of freedom increases significantly, which makes the framework less relevant for rare conditions.

This study has one undeniable limitation: each image was interpreted by a single human reader (an ophthalmologist), who was not obligated to annotate all visible conditions. Therefore, the quality of performance assessment could be improved. However, given the large number of conditions considered in this study, the relevance of the approach was clearly validated. Another limitation of this study is that alternatives to building blocks of the proposed framework, like PCA, Parzen-Rosenblatt or k-NN, were not explored.

In conclusion, we have presented the first study on the automatic detection of a large number of conditions in retinal images. A simple ML framework was proposed for this purpose. The results are highly encouraging and open new perspectives for ocular pathology screening. In particular, the trained detectors could be used to generate warnings when rare conditions are detected, both in traditional and automatic screening scenarios. We believe this will favor the adoption of automatic screening systems, which currently focus on the most frequent pathologies and ignore all others.

\bibliographystyle{elsarticle-harv}
\bibliography{spinoff}

\end{document}